# INTERACTIVE INFERENCE: A MULTI-AGENT MODEL OF COOPERATIVE JOINT ACTIONS

Domenico Maisto, Francesco Donnarumma, and Giovanni Pezzulo
Institute for Cognitive Sciences and Technologies, National Research Council, Rome, Italy

Contact: Giovanni Pezzulo (giovanni.pezzulo@istc.cnr.it)

Abstract—We advance a novel computational model of multi-agent, cooperative joint actions that is grounded in the cognitive framework of active inference. The model assumes that to solve a joint task, such as pressing together a red or blue button, two (or more) agents engage in a process of interactive inference. Each agent maintains probabilistic beliefs about the joint goal (e.g., should we press the red or blue button?) and updates them by observing the other agent's movements, while in turn selecting movements that make his own intentions legible and easy to infer by the other agent (i.e., sensorimotor communication). Over time, the interactive inference aligns both the beliefs and the behavioral strategies of the agents, hence ensuring the success of the joint action. We exemplify the functioning of the model in two simulations. The first simulation illustrates a "leaderless" joint action. It shows that when two agents lack a strong preference about their joint task goal, they jointly infer it by observing each other's movements. In turn, this helps the interactive alignment of their beliefs and behavioral strategies. The second simulation illustrates a "leader-follower" joint action. It shows that when one agent ("leader") knows the true joint goal, it uses sensorimotor communication to help the other agent ("follower") infer it, even if doing this requires selecting a more costly individual plan. These simulations illustrate that interactive inference supports successful multi-agent joint actions and reproduces key cognitive and behavioral dynamics of "leaderless" and "leader-follower" joint actions observed in human-human experiments. In sum, interactive inference provides a cognitively inspired, formal framework to realize cooperative joint actions and consensus in multi-agent systems.

# I. INTRODUCTION

A central challenge of multi-agent systems (MAS) is coordinating the actions of multiple autonomous agents in time and space, to accomplish cooperative tasks and achieve joint goals [1], [2]. Developing successful multi-agent systems requires addressing controllability challenges [3], [4] and dealing with synchronization control [5], formation control [6], task allocation [7] and consensus formation [8]–[10].

Research in cognitive science may provide guiding principles to address the above challenges, by identifying the cognitive strategies that groups of individuals use to successfully interact with each other and to make collective decisions [11]–[14]. An extensive body of research studied how two or more people coordinate their actions in time and space during cooperative (human-human) joint actions, such as when performing team sports, dancing or lifting something together [15], [16]. These studies have shown that successful joint actions engage various cognitive mechanisms, whose level of sophistication plausibly depends on task complexity. The simplest forms of coordination and imitation in pairs or groups of individuals, such as the joint execution of rhythmic patterns, might not require sophisticated cognitive processing, but could use simple mechanisms of behavioral synchronization – perhaps based on coupled dynamical systems, analogous to the synchronization of coupled pendulums [17]. However, more sophisticated types of joint actions go beyond the mere alignment of behavior. For example, some joint actions require making decisions together, e.g., the decision about where to place a table that we are lifting together. These sophisticated forms of joint actions and joint decisions might benefit from cognitive mechanisms for mutual prediction, mental state inference, sensorimotor communication and shared task representations [16], [18]. The cognitive mechanisms supporting joint action have been probed by numerous experiments [19]–[29], sometimes with the aid of conceptual [30], computational [31]–[39], and robotic [40]–[43] models. However, there is still a paucity of models that implement advanced cognitive abilities, such as the inference of others' plans and the alignment of task knowledge across group members, which have been identified in empirical studies of joint action. Furthermore, it is unclear whether and how it is possible to develop joint action models from first principles; for example, from the perspective of a generic inference or optimization scheme that unifies multiple cognitive mechanisms required for joint action.

We advance an innovative framework for cooperative joint action and consensus in multi-agent systems, inspired by the cognitive framework of active inference. Active inference is a normative theory that describes the brain as a prediction machine, which learns an internal (generative) model of the statistical regularities of the environment – including the statistics of social interactions – and uses it to generate predictions that guide perceptual processing and action planning [44]. Here, we use the predictive and inferential mechanisms of active inference to implement sophisticated forms of joint action in dyads of interacting agents. The model presented here formalizes joint action as a process of *inter*active inference based on shared task knowledge between the agents [2], [45]. We exemplify the functioning of the model in a "joint maze" task. In the task, two agents have to navigate in a maze, to reach and press together either a red or a blue button. Each agent has probabilistic beliefs about the joint task that the dyad is performing, which covers his own and the other agent's contributions (e.g., should we both press a red or a blue button?). Each agent continuously infers what the joint task is, based on his (stronger or weaker) prior belief and the observation of the other agent's movements towards one of the two buttons. Then, he selects an action (red or blue button press), in a way that simultaneously fulfills two objectives. The former, *pragmatic* objective consists in achieving the joint task efficiently (e.g., by following the shortest route to reach the to-be-pressed button). The latter, *epistemic* objective consists in shaping one's movements to help the other agent inferring what the joint goal is (e.g., by selecting a longer route easily associated to the goal of pressing the red button).

The next sections are organized as follows. First, we introduce the consensus problem (called a "joint maze") we will use throughout the paper to explain and validate our approach. Next, we illustrate the main tenets of the interactive inference model of joint action. Then, we present two simulations that illustrate the functioning of the *inter*active inference model. The first simulation shows that over time, the *inter*active inference aligns the joint task representations of the two agents and their behavior, as observed empirically in several joint action studies [18], [23], [46]–[49]. In turn, this form of "interactive alignment" (or "generalized synchrony") optimizes the performance of the dyad. The second simulation shows that when agents have asymmetric information about the joint task, the more knowledgeable agent (or "leader") systematically modifies his behavior, to reduce the uncertainty of the less knowledgeable agent (or "follower"), as observed empirically in studies of sensorimotor communication [16], [18]. This form of "social epistemic action" ensures the success of joint actions despite incomplete information. Finally, we discuss how our model of interactive inference could help better understand various facets of ("leaderless" and "leader-follower") human joint actions, by providing a coherent formal explanation of their dynamics at both brain and behavioral levels.

## II. PROBLEM FORMULATION AND SCENARIO

To illustrate the mechanisms of the *inter*active inference model, we focus on the consensus problem called "joint maze" task, which closely mimics the setting used in a previous human joint action study [50], see Fig. 1. In this task, two agents (represented as a grey hand and a white hand) have to "navigate" in a grid-like maze, reach the location in which the red or blue button is located, and then press it together. The task is completed successfully when both agents "press" the same button, whatever its color (unless stated otherwise).

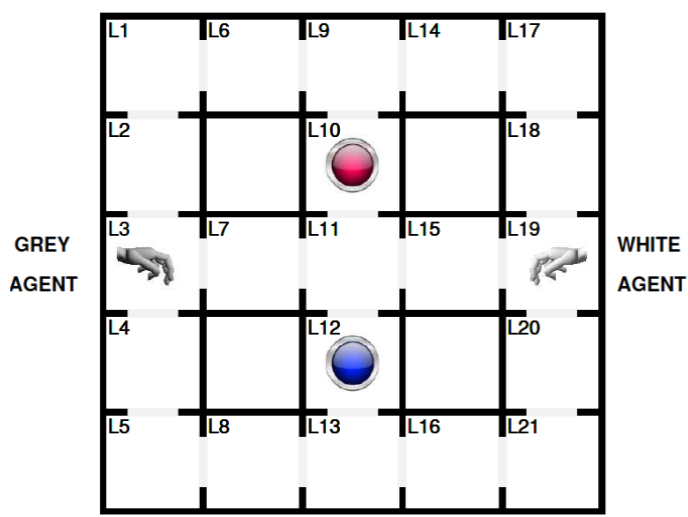

**Fig. 1.** Schematic illustration of the "joint maze" task. The two (grey and white) agents are represented as two hands. Their initial positions are L3 (grey) and L19 (white). Their possible goal locations are in blue (L12) and red (L10). The agents can navigate in the maze, by following the open paths, but cannot go through walls).

At the beginning of each simulation, each agent is equipped with some prior knowledge (or preference) about the goal of the task. This prior knowledge is represented as a probabilistic belief, i.e., a probability distribution over four possible task states; these are "both agents will press red", "both agents will press blue", "the white agent will press red and the grey agent will press blue" and "the white agent will press blue and the grey agent will press red". Importantly, in different simulations, the prior knowledge of the two agents can be congruent (if both assign the highest probability to the same state) or incongruent (if they assign the highest probability to different states); certain (if the probability mass is peaked in one state) or uncertain (if the probability mass is spread across all the states). This creates a variety of coordination problems, which span from easy (e.g., if the beliefs of the two agents are congruent and certain) to difficult problems (e.g., if the beliefs are incongruent or uncertain). Each simulation includes several trials, during which each agent follows a perception-action cycle. First, the agent receives an observation about his own position and the position of the other agent. Then, he updates his knowledge about the goal of the task (i.e., *joint task inference*) and forms a plan about how to reach it (i.e., *joint plan inference*). Finally, he makes one movement in the maze (by sampling it probabilistically from the plan that he formed). Then, a new perception-action cycle starts.

# III. METHODS

Here, we provide a brief introduction to the active inference framework for single agents (see [44] for details) and then we illustrate the novel, *inter*active inference model developed here to address multi-agent, cooperative joint actions.

### A. Active Inference

Active Inference agents implement perception and action planning through the minimization of variational free energy [44]. To minimize free energy, the agents use a generative model of the causes of their perceptions, which encodes the joint probability of the stochastic variables illustrated in Fig. 2, using the formalism of probabilistic graphical models [51].

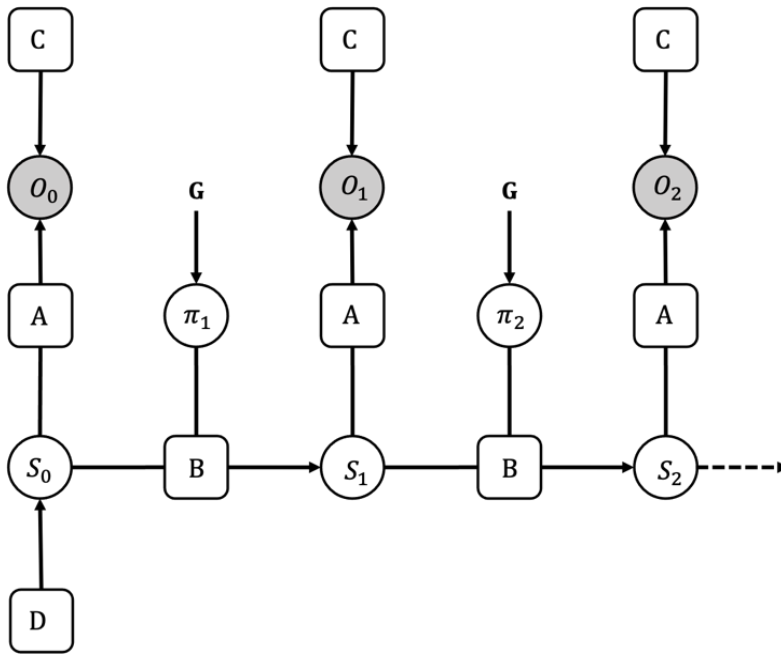

**Fig. 2.** Generative model of an active inference agent, unrolled in time. The circles denote stochastic variables; filled circles denote observed variables, whereas unfilled circles denote

variables that are not observed and need to be inferred. The squares indicate the categorical probability distributions that parameterize the generative model. Please see the main text for a detailed explanation of the variables.

The agent's generative model is defined as follows:

$$P(o_{0:T}, s_{0:T}, u_{1:T}, \gamma | \mathbf{\Theta}) = P(\gamma | \mathbf{\Theta}) P(\pi | \gamma, \mathbf{\Theta}) P(s_0 | \mathbf{\Theta}) \prod_{t=0}^{T} P(o_t | s_t, \mathbf{\Theta}) P(s_{t+1} | s_t, \pi_t, \mathbf{\Theta})$$

$$(1)$$

where $P(o_t | s_t, \mathbf{\Theta}) = \mathbf{A}$, $P(\pi_t | \gamma, \mathbf{\Theta}) = \sigma(\ln \mathbf{E} - \gamma \cdot \mathbf{G}(\pi_t) | \mathbf{\Theta})$, $P(s_{t+1} | s_t, \pi_t, \mathbf{\Theta}) = \mathbf{B}(u_t = \pi_t)$, $P(s_0 | \mathbf{\Theta}) = \mathbf{D}$, $P(\gamma | \mathbf{\Theta}) \sim \Gamma(\alpha, \beta)$.

The set $\mathbf{\Theta} = \{\mathbf{A}, \mathbf{B}, \mathbf{C}, \mathbf{D}, \alpha, \beta\}$ parameterizes the generative model. The (likelihood) matrix $\mathbf{A}$ encodes the relations between the observations O and the hidden causes of observations S. The (transition) matrix $\mathbf{B}$ defines how hidden states evolve over time $t$, as a function of a control state (action) $u_t$; note that a sequence of control states $u_1, u_2, \dots, u_t, \dots$ defines a policy $\pi_t$ (see below for a definition). The matrix $\mathbf{C}$ is an a-priori probability distribution over observations and encodes the agent's preferences or goals. The matrix $\mathbf{D}$ is the prior belief about the initial hidden state, before the agent receives any observation. Finally, $\gamma \in \mathbb{R}$ is a *precision* that regulates action selection and is sampled from a $\Gamma$ distribution with parameters $\alpha$ and $\beta$.

An active inference agent implements the perception-action loop by applying the above matrices to hidden states and observations. In this perspective, perception corresponds to estimating hidden states on the basis of observations and of previous hidden states. At the beginning of the simulation, the agent has access through $\mathbf{D}$ to an initial state estimate $S_0$ and receives an observation $O_0$ that permits refining the estimate by using the likelihood matrix $\mathbf{A}$. Then, for $t = 1, \dots, T$, the agent infers its current hidden state $S_t$ based on the observations previously collected and by considering the transitions determined by the control state $u_t$, as specified in $\mathbf{B}$. Specifically, active inference uses an approximate posterior over (past, present and future) hidden states and parameters $(s_{0:T}, u_{1:T}, \gamma)$. Assuming a mean field approximation, it can be described as:

$$Q(s_{0:T}, u_{1:T}, \gamma) = Q(\pi) Q(\gamma) \prod_{t=0}^{T} Q(s_t | \pi_t) \qquad (2)$$

where the sufficient statistics are encoded by the expectations $\boldsymbol{\mu} = (\tilde{\boldsymbol{s}}^{\pi}, \boldsymbol{\pi}, \boldsymbol{\gamma})$, with $\tilde{\boldsymbol{s}}^{\pi} = \boldsymbol{s}_0^{\pi}, \dots, \boldsymbol{s}_T^{\pi}$. Following a variational approach, the distribution in Eq. (2) best approximates the posterior when its sufficient statistics $\boldsymbol{\mu}$ minimise the free energy of the generative model, see [44]. This condition holds when the sufficient statistics are:

$$\boldsymbol{s}_t^{\pi} \approx \sigma(\ln \mathbf{A} \cdot o_t + \ln(\mathbf{B}(\pi_{t-1}) \cdot \boldsymbol{s}_{t-1}^{\pi})) \qquad (3.a)$$

$$\boldsymbol{\pi} = \sigma(\ln \mathbf{E} - \boldsymbol{\gamma} \cdot \mathbf{G}(\pi_t)) \qquad (3.b)$$

$$\boldsymbol{\gamma} = \frac{\alpha}{\beta - \mathbf{G}(\boldsymbol{\pi})} \qquad (3.c)$$

where the symbol "$\cdot$" denotes the inner product, defined as $\mathbf{A} \cdot \mathbf{B} = \mathbf{A}^T \mathbf{B}$, with the two arbitrary matrices $\mathbf{A}$ and $\mathbf{B}$.

Action selection is operated by selecting the policy (i.e., sequence of control states $u_1, u_2, \dots, u_t$) that is expected to minimize free energy the most in the future. The policy distribution $\boldsymbol{\pi}$ is expressed in (3.b); the term $\sigma(\cdot)$ is a softmax function, $\mathbf{E}$ encodes a prior over the policies (reflecting habitual components of action selection), **G** is the expected free energy of the policies (reflecting goal-directed components of action selection) and γ is a precision term that encodes the confidence of beliefs about **G**.

The expected free energy (EFE) **G**($\pi_t$) of each policy $\pi_t$ is defined as:

$$\mathbf{G}(\pi_t) \approx \sum_{\tau=t+1}^{T} -D_{\text{KL}}[Q(o_\tau|\pi)||P(o_\tau)] - \mathbb{E}_{\tilde{Q}}\big[H[P(o_\tau|s_\tau)]\big]$$

(4)

where $D_{\text{KL}}[\cdot || \cdot]$ and $H[\cdot]$ are, respectively, the Kullback-Leibler divergence and the Shannon entropy, $\tilde{Q} = Q(o_\tau, s_\tau|\pi) \triangleq P(o_\tau|s_\tau)Q(s_\tau|\pi)$ is the predicted posterior distribution, $Q(o_\tau|\pi) = \sum_{s_\tau} Q(o_\tau, s_\tau|\pi)$ is the predicted outcome, $P(o_\tau)$ is a categorical distribution representing the preferred outcome and encoded by **C**, and $P(o_\tau|s_\tau)$ is the likelihood of the generative model encoded by the matrix **A**.

The expected free energy (EFE) can be used as a quality score for the policies and has two terms. The first term of (4) is the Kullback-Leibler divergence between the (approximate) posterior and prior over the outcomes and it constitutes the pragmatic (or utility-maximizing) component of the quality score. This term favours the policies that entail low risk and minimise the difference between predicted ($Q(o_\tau|\pi)$) and preferred ($P(o_\tau) \equiv$ **C**) future outcomes. The second term of (4) is the expected entropy under the posterior over hidden states and it represents the epistemic component of the quality score. This term favours policies that lead to states that diminish the uncertainty future outcomes $H\big[P\big(o_\tau|s_\tau\big)\big]$.

After scoring all the policies using EFE, action selection is performed by drawing over the action posterior expectations derived from the sufficient statistic $\boldsymbol{\pi}$ computed via (3). Then, the selected action is executed, the agent receives a novel observation and the perception-action cycle starts again. See [44] for more details.

### B. Multi-Agent Active Inference

Here, we extend the active inference framework outlined above to a multi-agent setting [1], in which multiple agents (here, two) perform a joint task consisting in navigating in the "joint maze" of Fig 1 to reach simultaneously either the red or the blue location. The "joint maze" of Fig 1 includes 21 locations {L1, L2, … , L21}. Two agents, "grey" ("i") and "white" ("j"), start from the locations L3 and L19 and their goal is to reach either the red (L10) or the blue (L12) goal locations. To reach the goal, the agents can choose between 24 action sequences or policies π (see Fig. S1 for their full list), which can be divided into two main types: those that follow the shorter paths that pass through the central corridor or longer paths that go through the maze perimeter. The shorter paths of the grey agent to reach the red and blue goal locations are (L3, L7, L11, L10) and (L3, L7, L11, L12), respectively. The longer paths of the grey agent to reach the red and blue locations are (L3, L2, L1, L6, L9, L10) and (L3, L4, L5, L8, L13, L12), respectively. The shorter paths of the white agent to reach the red and blue goal locations are(L19, L15, L11, L10) and (L19, L15, L11, L12), respectively. The longer paths of the grey agent to reach the red and blue locations are (L19, L18, L17, L14, L9, L10) and (L19, L20, L21, L16, L13, L12), respectively. Below, we will call the first type of policies that go through the shorter paths "pragmatic policies" and the second type of policies that go through the longer paths "social epistemic policies".

Each agent has a separate generative model, whose structure was shown in Fig 2. In simulation 1, the two agents are equipped with identical generative models, except for a different estimate of their starting positions, L3 or L19. In simulation 2, there are some differences in the **A** and **D** tensors of the two agents (see below), reflecting the fact that the white agent (the "leader") knows the joint task to be performed, whereas the grey agent (the "follower") does not.

When the two generative models are considered together, they can be defined as $\langle S^i, O^i, U^i, \Theta^i \rangle$, with $i = 1, \ldots, N$, where N is the number of agents (see Supplementary Fig. S8C). Here, we assume that $N = 2$, but it is possible to generalise the model to a larger number of agents. The hidden states $S^i = S_1^i \otimes S_2^i \otimes S_3^i$ are obtained as a tensorial product of three vectors (note that unlike the usual algebraic notation for tensors, the subscripts and superscripts do not indicate covariance or contravariance). The three vectors are: the position of the agent $S_1^i$, the position of the other agent $S_2^i$, and the joint goal context $S_3^i$, that is the agent's belief about which joint goal the agents could achieve (see Fig. 3B). In this study, where the potential goals are two, there are four possible combinations: blue-blue, blue-red, red-blue, and red-red; hence, the size of the hidden state is $21 \times 21 \times 4 = 1764$.

The observations $O^i = O_1^i \otimes O_2^i \otimes O_3^i \otimes O_4^i$ consist of the tensorial product between the observed agents' positions $O_1^i, O_2^i$, the observed joint goal $O_3^i$, and the associated utilities $O_4^i$ (see Supplementary Fig. S8B). The first three vectors encode the observations that correspond to the three sets of hidden states described above. The final vector, the outcome utility, includes three observations that correspond to the possible task outcomes: negative, neutral or positive / rewarding. The control states $U^i = U_1^i \otimes U_2^i$ denote the joint actions available to the agents. Note that in this simulation, each agent has beliefs about his own and the other agent's control states, even if of course he can only execute his own actions

The set of tensors $\Theta^i = \{A^i, B^i, C^i, D^i\}$ define the structure of the generative model. The tensors $D^i = D_1^i \otimes D_2^i \otimes D_3^i$ and $C^i = C_1^i \otimes C_2^i \otimes C_3^i \otimes C_4^i$ encode the priors of the hidden states and the observations, respectively. The former factor reflects the agent's prior knowledge about its initial state. We assume that each agent knows his own and the other agent's initial positions ($D_1^i$ and $D_2^i$ are deterministic), but the belief $D_3^i$ about which goal the other expects is uncertain and adjustable as a simulation parameter. Besides, $D_3^i$ depends on the agent's role, "leader" or "follower". The leader knows the joint task goal and hence it splits the probability mass of $D_3^i$ equally between blue-blue and blue-red (or between red-red and red-blue). Conversely, the follower only knows that, in order to succeed, both agents have to achieve the same goal. Hence, he splits the probability mass of $D_3^i$ equally between blue-blue and red-red. The factor $C_4^i$ (i.e., a prior over observations that incentivizes preferred outcomes) depends on the agent's role, in the same way as $D_3^i$. The (likelihood) mapping between $S^i$ and $O^i$ is specified through the tensor $A^i$, defined as the tensorial product $A^i = \otimes_k A_k^i$, $k = 1, \ldots, 4$, where each $A_k^i$, one for each different factor of $O^i$, is a 4-order tensor defined on the hidden states. The first factor $A_1^i$ is an identity tensor that maps the hidden states that represent the agent's positions in the maze into their corresponding observations.

The tensor $A_2^i$ maps the hidden states of the agent into the observations $O_2$ that are relative to the other agent's location in the maze. This (likelihood) mapping is regulated by a modulation factor $\delta_{s_1 s_2}^D$, which depends on the salience of the positions of the two agents ($s_1$ and $s_2$) and the beliefs about the goals $D_3^i$, i.e., $p(o_2|s_1, s_2) \propto \delta_{s_1 s_2}^D$. Intuitively, when the agents' positions provide diagnostic information about their joint task goal (e.g., the agents are close to one of the goal locations and far from the other), they are salient and their corresponding observations are not attenuated. Rather, when the agents' positions are uninformative (e.g., the agents are equidistant from the two goal locations), they are not salient and their corresponding observations are attenuated.

Note that the tensor $A_2^i$ can be intuitively considered as a kind of "salience map", which unlike previous active inference studies [44], is not fixed but varies as a function of the agent's beliefs about the joint task goal, hence prioritizing information processing in task-dependent ways [52]. In this setting, the salience can be interpreted as a form of attention modulation or gain-control (or even situation awareness [53]), in the sense that an agent receives noisy observations about the location of the other agent (specified in the state vector $s_2$) and the salience of the location determines the amount of endogenous (i.e., agent-dependent) noise in the observation process.

Mechanistically, we calculate the tensor $A_2^i$ in two steps. The first step consists in calculating the salience of the position $s_i$ of each agent $i$ as $\Delta v^i = \left| v_b^D(s_i) - v_r^D(s_i) \right|$, or the (absolute) difference between the salience of the position with respect to the blue goal $v_b^D(s_i)$ and the red goal $v_r^D(s_i)$. In turn, the salience with respect to one of the goals (here, the blue goal) is calculated as follows:

$$v_b^D(s_1) = \left(1 - \left(\max(D_3^i) - 0.5\right)\right) \cdot \left(\frac{d(s_1, L10)}{d(s_1, L10) + d(s_1, L12)}\right) \qquad (5)$$

According to Equation (5), the salience of the location $s_1$ with respect to the blue goal is the probability that an agent in that state is pursuing that goal. For simplicity, Equation (5) assumes that the closer an agent to the blue goal location (L12 in Fig. 1), the higher the probability that the agent is pursuing it. More specifically, the salience mass function $v_b^D$ relative to the position $s_1$ is inversely proportional to the Euclidean distance between $s_1$ and the goal (the second factor of Equation (5)). Furthermore, the salience depends on $D_3^i$ (the first factor of Equation (5)): the higher the $D_3^i$'s mode – i.e., the less entropic $D_3^i$ is – the smaller the amplitude of the salience. Note that we could have designed a more sophisticated salience model that considers the agent's direction of movement rather than, or in addition to, agent-goal distance; but in this simple setup, agent-goal distance is sufficient. Note also that we can define the salience $v_r^D(s_1)$ of the state $s_1$ with respect to the red goal by swapping L10 and L12 in Equation (5).

The second step to calculate $A_2^i$ consists in using the saliences calculated in the first step to calculate the modulation factor $\delta_{s_1 s_2}^D = \mathrm{sig}(\Delta v^1 \cdot \Delta v^2)$, where $\mathrm{sig}(x) = 1/(1 + h \cdot e^{-kx})$ is the parametric logistic function. For the purpose of our simulations, we assume that the modulation ranges in the interval $(0.75, 1)$, which we obtain by fixing the parameters $h = 10$, $k = 4$. Therefore, when the agents' positions unambiguously reveal the joint task goal they are pursuing, there is no modulation ($\delta_{s_1 s_2} = 1$). Rather, the amount of modulation increases by a maximum of 25% ($\delta_{s_1 s_2} = 0.75$) when the agents' positions provide poor information about their joint task goal.

The tensor $A_3^i$ encodes the joint position (and salience) of the two agents. Given the hidden state $s = s_1 \otimes s_2 \otimes s_3$, by assuming the saliences $v_b^D(s_1)$ and $v_b^D(s_2)$ as independent, it is possible to assume that $A_{3\,1,2,3}^i \equiv v_{s_3}^D(s_1, s_2) = v_{s_3}^D(s_1) \cdot v_{s_3}^D(s_2)$. Hence, we can associate with the $A_3^i$ components the corresponding joint saliences computed as the hidden state $s = s_1 \otimes s_2 \otimes s_3$ changes. Supplementary Fig. S9 shows various salience matrices of $A_3^i$ organized in a table, in which where the rows indicate joint goal context $s_3$ and the columns correspond to different values of the mode of the initial belief about the task goal i.e., $\max(D_3^i)$). Each matrix is a grid of size $21 \times 21$, where the rows and the columns represent the positions $s_1$ and $s_2$, respectively, the colours of the cells correspond to the values of the joint salience $v_{s_3}^D(s_1, s_2)$. Supplementary Fig. S9 shows that the matrices in the first column ($\max(D_3^i) = 0.5$) encode saliencies that are more peaked around the joint task goals. The matrices shown in the next two columns encode gradually more uniform and lower-valued saliencies – up until only the goal locations are salient. Note that in the control simulation (described in the section on Simulation 1) in which we prevent interactive inference to take place, we replace the $A_3^i$ tensor shown in Supplementary Fig. S9 with a uniform tensor that does not allow inferring the goal of the other agent from its position.

The tensor $A_4^i$ is responsible for modelling the relationship between hidden states and outcomes, which can be positive, neutral or negative. $A_4^i$ is a deterministic sparse tensor. For any hidden state s corresponding to a joint position $s_1 \otimes s_2$ that does not include any goal location, $A_4^i$ gives a "neutral" outcome. The definition of "positive" and "negative" outcomes varies depending on the joint task goal and the (leader or follower) roles of the agents. In Simulation 1, where both agents are treated as "followers", the outcome is positive if the both agents are in the same goal location (e.g., both are in L10 or both are in L12). Rather, the outcome is negative if only one agent is on a goal location or if the two agents are in two distinct goal locations. In Simulation 2, one of the two agents (the "follower") has the same tensor $A_4^i$ as in Simulation 1. Rather, the other agent (the "leader") receives a positive outcome if both agents are in the correct goal location (e.g., L10 if the joint task goal is "red, red") but a negative outcome if at least one agent is in the incorrect goal location (e.g., L12 if the joint task goal is "red, red").

The tensor B encodes a deterministic mapping between hidden states, given the control state u. Note that here the control state u corresponds to a joint action, not to the action of a single agent; hence it is specified as the tensorial product between the vector of the five possible movements of one agent ('up', 'down', 'left', 'right', and 'wait') and the vector of the same five movements of the other agent. The tensor B describes how the spatial locations $s_1$ and $s_2$ of the agents change as a function of the joint actions $u^i = u_1^i \otimes u_2^i$, such as 'up-up', 'up-down', 'up-left', etc. Note that the transitions regard exclusively the spatial locations. The transitions between joint task goals are not modelled explicitly, but the agents can infer that the joint goal changed on the basis of their observations.

The action-perception cycle of the multi-agent active inference model is the same as the single-agent active inference (see Fig. 2), except that the two agents exchange observations between them. Specifically, agent "i" receives from the agent "j" the outcome vectors $O_1^j$ and the action $u_1^j$ and vice versa (for simplicity, we allow the agents to send actions to each other; a slightly more complex generative model could have included as inputs observations about others' actions rather than directly their actions). Each agent then uses this information (along with the observations $O_2^i$ and the actions $u_2^i$ that they have computed) to update his beliefs about the hidden states and control states and then to select a course of actions or policy π.

The two key mechanisms of the model are *Task goal inference* and *Plan inference. Task goal inference* corresponds to inferring what the goal of the joint task is, i.e., updating the belief about the four possible task goals ("blue, blue", "blue, red", "red, blue", "red, red"). As the task goal is specified at the level of the dyad, in order to infer it, each agent needs to consider both his prior knowledge about the task goal and the movements of the other agent, which are informative about the other agent's task knowledge. Specifically, the joint task goal inference follows a principle of rational action; namely, the expectation that the other agent will act efficiently to achieve his goals [54]. Put simply, if an agent observes the other agent moving towards the red (or blue) goal, he updates his belief about the joint task goal, by increasing the probability that the goal is red (or blue). Furthermore, at the end of each trial, both agents receive feedback about success ("win" observation) or failure ("lose" observation) and update their beliefs about the task goal. Please see the Methods section for more details.

*Plan inference* corresponds to inferring the course of action (or plan) that maximizes task success, on the basis of the inferred joint task. In this model, each agent infers both his own and the other agent's plan – although, of course, he can only execute his own plan. The inference about one's own and the other agent's plans needs to consider the utility of following different routes (which privileges the shortest route) and the uncertainty about the goal (which prompts "epistemic" behavior and the selection of informative routes). The balance between utilitarian and epistemic components of planning will become important in Simulation 2, see later.

A key thing to notice is that the perception-action cycles of the two agents – and their inferential processes – are mutually interdependent, as the movements of one agent determine the observations of the other agent at the next time step. Our simulations will show that this interactive inference naturally leads to the alignment of beliefs states and behavioral patterns of the two agents, analogous to the synchronization of neuronal activity and kinematics in socially interacting dyads [15], [29], [48]. Furthermore, the simulations will show that "social epistemic actions" that aim at reducing the uncertainty of the other agent increase the alignment and task success, especially in tasks with asymmetric knowledge.

# IV. SIMULATIONS OF INTERACTIVE INFERENCE

We present two simulations of interactive inference. The first simulation illustrates the case of two agents that start from various (uncertain) prior beliefs about the task goal. This simulation illustrates that over time, interactive inference produces a gradual alignment of both belief states and behavior, which permits the agents to successfully complete the task (most of the times). The second simulation illustrates the case of two agents that initially have asymmetric task knowledge: one of them (the leader) has a certain prior belief about the task goal, whereas the other (follower) has an uncertain belief. This simulation illustrates sensorimotor communication – and the importance for the leader to select (epistemic) actions that reduce the follower's uncertainty, in order to complete the task successfully.

## A. Simulation 1: leaderless interaction

The goal of Simulation 1 is testing whether and how interactive alignment favors the alignment of behavior and belief states of two agents engaged in the "joint maze" task. This simulation comprises 100 trials. For each trial, two identical agents (apart for their prior beliefs about the task goal, see later) start from two opposite locations of the "joint maze": the grey agent starts in location L3 and the white agent starts in location L19. They can move one step at a time, or wait (in which case, they remain the same place), until they reach one of the locations that include colored goals (red in L10, blue in L12). There are multiple sequences of actions (aka "policies") that each agent can take to reach the goal locations, which correspond to shorter or longer paths, with or without "wait" actions, etc. The 25 policies used in the simulation are specified in the Methods section. What is most important here is that irrespective of the selected policy, a trial is only successful if both agents reach the same goal / button location, red (L19) or blue (L12). Specifically, if at the end of the trial both agents are in the red (L10) or the blue (L12) button location, then the trial is successful and the agents receive the preferred observation ("win"). Otherwise, if the two agents fail to reach the same button location (e.g., one is in L10 and the other is in L12), the trial is unsuccessful and the agents receive an undesirable observation ("lose").

Fig. 3 shows the results of one example simulation, in which the agents start with the same prior belief about the task goal. This uncertain belief assigns 0.5 to "both agents will press red" (in short, "red, red"), 0.5 to "both agents will press blue" (in short, "blue, blue") and zero to the two other possible states ("red, blue: the white agent will press red and the grey agent will press blue" and "blue, red: the white agent will press blue and the grey agent will press red").

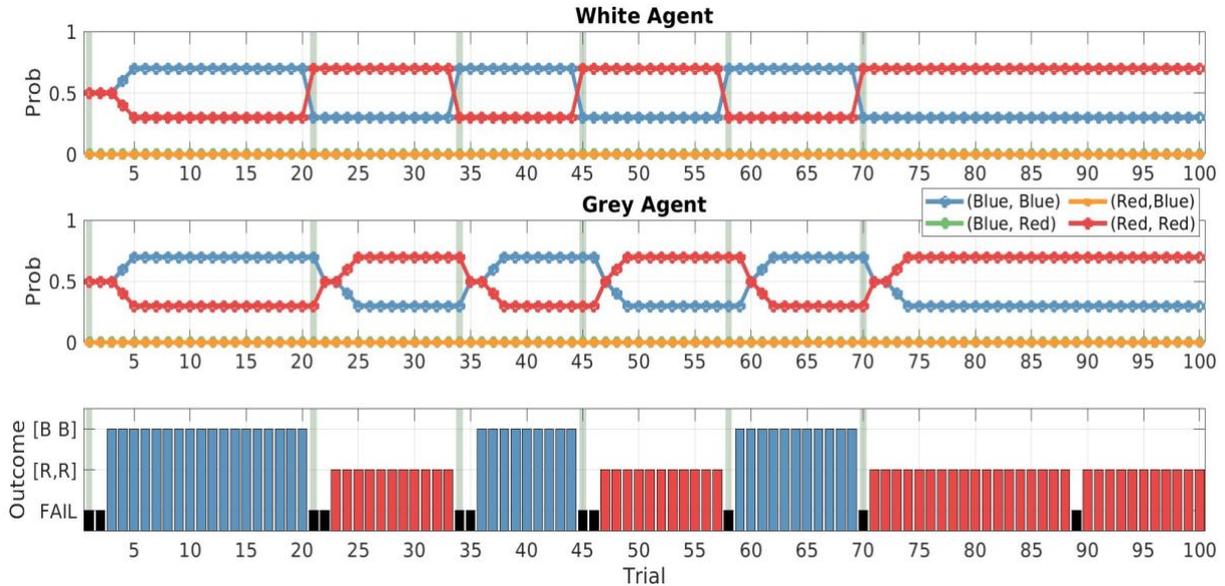

**Fig. 3.** Results of Simulation 1. The first two panels show the prior beliefs of the white (first panel) and grey (second panel) at the beginning of each trial. The vertical bars indicate moments in which we manually change the mind of the "white agent" and set his belief about the joint task goal is set to 1 for "blue, blue" (if its prior belief assigned higher probability to "red, red") or "red, red" (if its prior belief assigned higher probability to "blue, blue"). The third panel shows the outcome of the trials. These include successful trials in which the agents press the blue button (blue bars) or the red button (red bars) and failures (black bars).

The first two panels of Fig. 3 show the prior beliefs of the white and grey agents, respectively, at the beginning of each trial, from 1 to 100. The agents' prior beliefs for the first trial are set manually, as discussed above. The agents' prior beliefs for the subsequent trials are simply the posterior beliefs at the end of the previous trials, as usual in Bayesian inference, but multiplied by a fixed (volatility) factor. This ensures that the prior probability of "red, red" or "blue, blue" cannot be higher than 0.7. The reason for introducing the fixed factor is that in many trials, the posterior beliefs reach the value of 1 for "red, red" or for "blue, blue", indicating that the agent is fully sure about the shared task goal. If this posterior value were used as the prior value for subsequent trials, there would be little place for variability in behavior and changes of mind. Introducing the fixed factor amounts to assuming that the agents are not fully sure that that joint task goal would remain the same across trials – or in other words, believe that the environment has some volatility. Note that from time to time (vertical bars) we manually "change the mind" of agent 1 from "red, red" to "blue, blue" or vice versa – to introduce some variability in the simulation.

The third panel of Fig. 3 shows whether the agents completed successfully the trial by pressing the same button (the blue bars indicate that both pressed the blue button, whereas the red bars indicate that both pressed the red button) or unsuccessfully (black bars). Finally, the green vertical bars show trials in which the white agent "changes mind" about the goal (e.g., from "blue, blue" to "red, red" or vice-versa). Following a "change of mind", the dyad usually requires one or a few trials before re-aligning on the new joint task goal.

Fig. 3 shows that the two agents end up the trials with aligned belief states most of the times, except in the first trials (in which they started with uncertain beliefs) and immediately after the changes of mind (vertical bars). Furthermore, the two agents are successfully during most of the trials in which their beliefs are aligned and unsuccessful when their beliefs are not aligned. As shown in Fig. 3, the errors occur in the very first trials, immediately after the grey agent changes mind and in one trial afterwards. The errors on the first trials may occur because the agents are uncertain about what to do and they assign the same (expected free energy) "score" to the two policies that go straight to the red button and the blue button; see the Methods section for an explanation of (expected free energy) and Fig. S1 for an illustration of the expected free energy of the policies of the white agent at the beginning of the first trials. When the two agents are very uncertain, there are two possible behaviors. First, both agents may select their task goals randomly, which might or might not result in an error (see Fig. S2 for an illustration of the results of 100 replications of the same experiment, without changes of mind). Second, one agent might simply follow the other and be successful. This "follower effect" is particularly apparent when the agents' prior beliefs are weak, as in the first trials. Rather, in trials in which the agents' prior beliefs are strong, such as after a change of mind, they do not simply follow one another, but try to fulfill their prior belief – and this explains why we observe several errors after only one of the agents changes mind. These examples illustrate that it is the strength (or the precision) of the beliefs about the joint task goal that determines whether or not an imitative response takes place; see also [40] for a robotic demonstration of the importance of prior beliefs in enabling imitative responses. Finally, note that some errors can occur randomly, with low probability, since action selection is stochastic.

To better quantify the interactive alignment of belief states between the agents across trials and its effects on performance, we executed 100 runs of Simulation 1 and plotted a measure of the belief alignment of the dyad – the KL divergence between the beliefs about task goals – and their performance, during the first 15 trials; see Fig. 4. Fig. 4A shows that the KL distance between the prior beliefs of the two agents is initially low, because they both start the same by design. While the beliefs are apparently aligned, the alignment regards an uncertain state – and this is why the performance is initially low (Fig. 4B). During the interaction, the KL divergence initially increases, as the agents consider different hypotheses about the joint task goal, but then it rapidly decreases when the agents settle on the same joint task goal ("red, red" or "blue, blue") – and at that point, the performance of the dyad (Fig. 4B) is nearly perfect.

Note that in this simulation the initial choice of a particular joint task goal ("red, red" or "blue, blue") is random, but its persistence across trials depends on a process of interactive belief alignment between the agents. The alignment of behavior and of beliefs about task goals might occur in two ways. First, it might occur thanks to interactive inference within trials: namely, because each agent monitors the movements of the other agent and uses this information to update his estimate about the joint task goal and the other agent's plan, following a principle of rational action (i.e., the expectation that the other agent will act efficiently to achieve his goals [54]). Second, the alignment might occur because at the end of each trial, the agents receive a feedback about their success ("win" observation) or failure ("lose" observation) and use this feedback to update their beliefs. Second, alignment might be the byproduct of a standard reinforcement learning approach to learn behavior by trial and error, without interactive inference within trials.

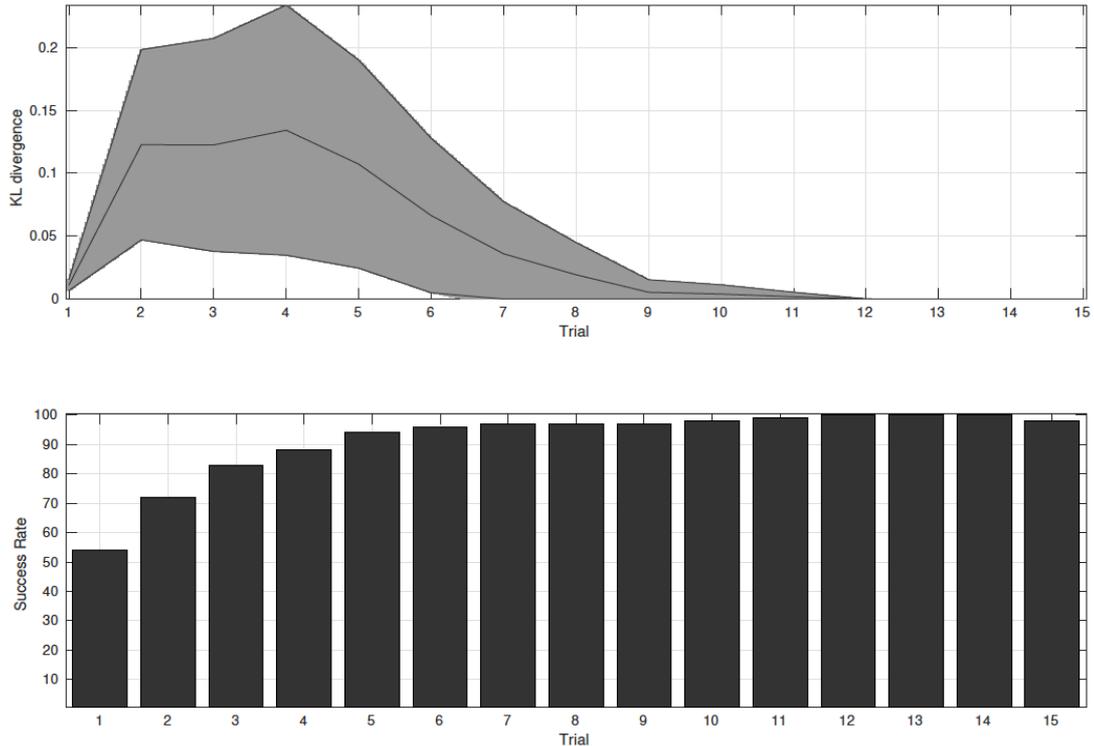

**Fig. 4.** Average results of 100 runs, with the same parameters as Simulation 1, for 15 trials. The top panel shows a measure of belief (dis)alignment of the agents: the KL divergence between their beliefs about task goals. The mean value is in black and the standard deviation of the mean is in grey. The bottom panel shows a histogram of mean success rate. See the main text for explanation

To understand whether the first mechanism based on interactive inference is actually useful for alignment and task success, we replicated the same experiment, but by preventing interactive inference to take place. We did this by removing any useful information from the likelihood matrix that maps the others' positions into task goals (i.e., by making the $A_3^i$ tensor uniform; see the Methods section for details). This control simulation shows that without interactive inference, the performance decreases drastically and there is little alignment: the agents keep switching between red and blue goals and their beliefs do not become increasingly aligned over time; see Supplementary Fig. S3 and S4. This control simulation shows that despite the task could be addressed using (reinforcement-based) feedback from successes and failures, the interactive inference is key to achieve alignment. Increasing the weight assigned to feedback information could potentially increase success rate and alignment, but this does not seem necessary when interactive inference is in place.

In sum, Simulation 1 shows that two agents that engage in interactive inference can align both their beliefs about the joint task goal and their plans to achieve the joint task goal, forming shared task knowledge [55]–[57]. The alignment at both the belief and behavioral levels is made possible by a process of interactive, reciprocal inference of goals and plans. The two agents initially have weak beliefs about the goal identity and therefore they can "follow each other" until they settle on some joint goal – and successively stick to it.

### B. Simulation 2: asymmetric leader-follower interaction

The goal of Simulation 2 is testing the emergence of "leader-follower" dynamics observed in human studies using the "joint maze" setup [50] and other related studies in which the agents have asymmetric preferences (or information) about the joint task goal [24], [58]–[60]. This simulation is similar to Simulation, 1, but the two agents differ in their prior beliefs about the task goal [24], [50], [58]–[60]. Specifically, the white agent (the "leader") knows the task to be accomplished – for example, "red, red" – whereas the grey agent (the "follower") does not. In other words, while in Simulation 1 both agents had initially weak beliefs (or preferences) about the joint goal and can be therefore considered two "followers", in Simulation 2 one of the two agents is a "leader" and has a strong initial preference about the joint task goal.

The generative model of the follower is identical to the one used in Simulation 1, whereas the generative model of the leader differs from it in two ways. First, the (likelihood) tensor $A_4^i$ of the white agent reflects his knowledge of the true task contingencies; namely, that the preferred "win" observation can only be obtained by achieving the joint task "red, red", but not "blue, blue" (or the opposite when the true task goal is "blue, blue"). Furthermore, the prior belief of the white agent is 0.5 for "red, red", 0.5 for "red, blue" and 0 for the two other joint goals (or the opposite when the true task goal is "blue, blue"). Note this prior belief makes the white agent resistant to changes of mind about the task goal.

Several studies [24], [50], [58]–[60] showed that when leaders and followers have asymmetric information, the leaders modify their movement kinematics to "signal" their intentions and reduce the uncertainty of the followers [61], [62]. For example, consider that in the scenario of Fig. 1 the leader (white agent) has a choice between two kinds of action sequences or policies to reach the red goal location. The first, "pragmatic policies" follow the shortest and hence most efficient path to the goal: L15, L11 and L10. However, if the leader selects the pragmatic policy, he does not offer any cue to the follower about the joint task goal, until the last action (to L10). This is because passing through L15 and L11 is equally likely if the intended goals are red or blue and hence does not provide diagnostic information about the goal location. The second, "social epistemic policies" follow the route through L18, L17, L14, L9 and L10, which despite being longer, provides to the follower early information to the intended goal location. This is because passing through L18, L17, L14, L9 is rational only if the goal is the red button – and hence it provides diagnostic evidence that the to-be-pressed button is red. The above studies [24], [50], [58]– [60] show that leaders often select "social epistemic policies": they sacrifice efficiency to reduce the follower's uncertainty.

The trade-off between pragmatic and epistemic components of policy selection is automatic in active inference, because the expected free energy functional used in active inference to score policies includes two components: a "pragmatic component" that maximizes utility and prioritizes the shortest paths to the goal and an "epistemic component" that minimizes uncertainty (see the Methods). We therefore expected the leaders to select "social epistemic policies" most often when the followers were uncertain – and select "pragmatic policies" when uncertainty was reduced.

The results of an example leader-follower simulation lasting 30 trials are shown in Fig. 5. The first and third panels of Fig. 5 show the prior beliefs of the leader (white agent) and the follower (grey agent), respectively, at the beginning of each trial. These are largely aligned, except in the very first trials. The second and fourth panels show the policies selected by the leader and the follower, respectively. As discussed above, we divided policies into two categories: "pragmatic policies" (S: shorter red bars) that follow the shortest path to the goal and "social epistemic policies" (L: longer blue bars) that follow longer but more informative paths. The second panel of Fig. 5 shows that the leader tends to select "social epistemic policies" in the first trials, to reduce the follower's uncertainty (see also Supplementary Fig. S5 for a visualization of the expected free energy of the leader's policies). Rather, the follower has no benefit from selecting epistemic policies and indeed selects pragmatic policies across almost all the trials. Finally, the bottom panel of Fig. 5 shows that in all but the first two trials (short black bars), the agents successfully achieve the "red, red" joint goal (long red bars).

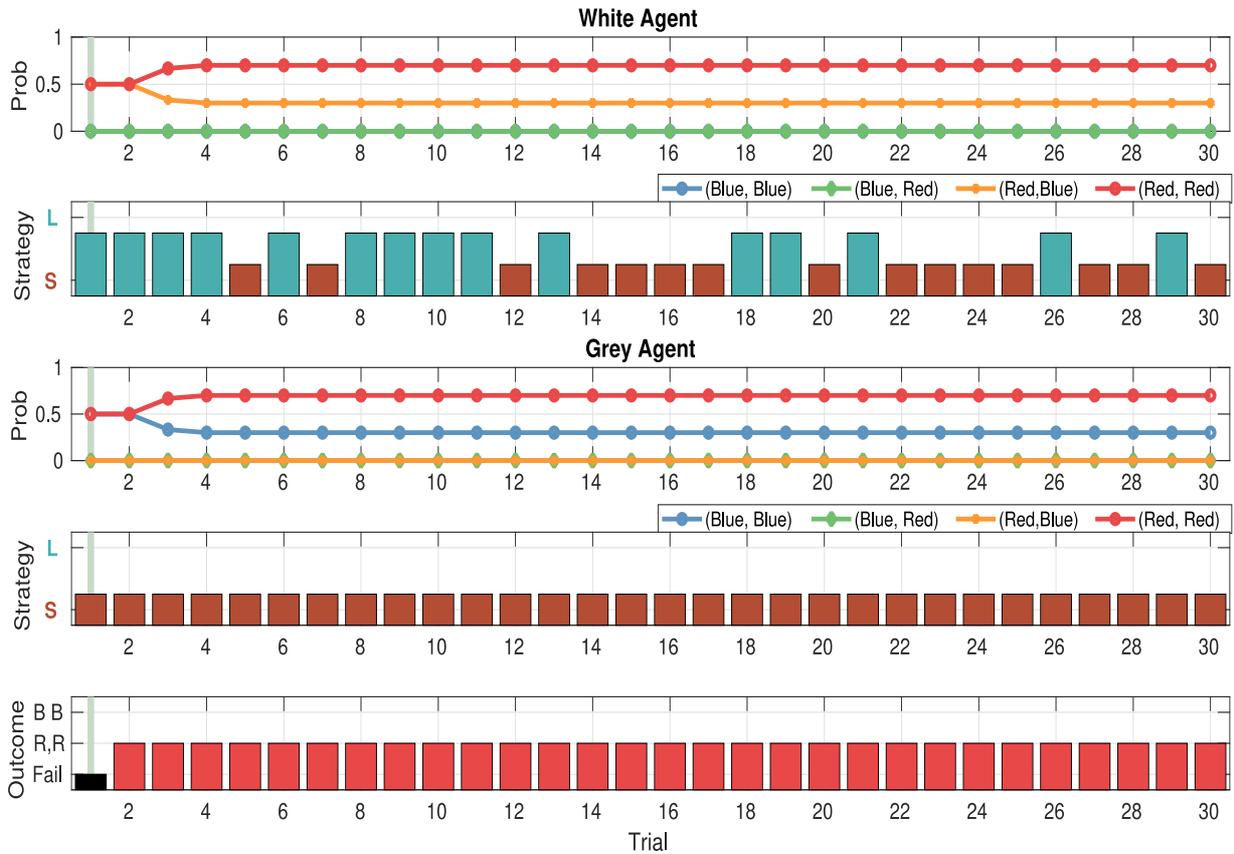

**Fig. 5.** Results of Simulation 2: example of leader-follower dyadic interaction, for 30 trials. The first two panels show the prior beliefs of the leader at the end of each trial and the policy he selects (shorter red bar: pragmatic policy that follows the shortest path to the goal; longer blue bar: social epistemic policy that follows the longer but more informative path to the goal). The third and fourth panels show the prior beliefs of the follower at the end of each trial and the policy he selects. The fifth panel shows the outcome of the trials. These include successful trials in which the agents press the red button (red bars) and failures (black bars).

Fig. 6 shows the results of 100 repetitions of the same simulation (see also Supplementary Fig. S6). The first panel shows that the beliefs of the agents, measured as the KL divergence between their prior beliefs about the task goal, aligns over time. The second panel shows that the average performance of the dyads, measured as the number of times they select the correct "red, red" goal, increases over time. The third panel shows the percentage of "social epistemic policies" selected by the leaders. Initially, the leaders have a strong tendency to select "social epistemic policies", but this tendency decreases significantly across trials, as the followers become increasingly certain about the joint task goal – which is in agreement with empirical evidence [24], [50], [58]–[60]. This result emerges because in the EFE that active inference uses to score policies, the decrease of uncertainty lowers the epistemic value of policies, hence lowering the probability that they will be selected [63].

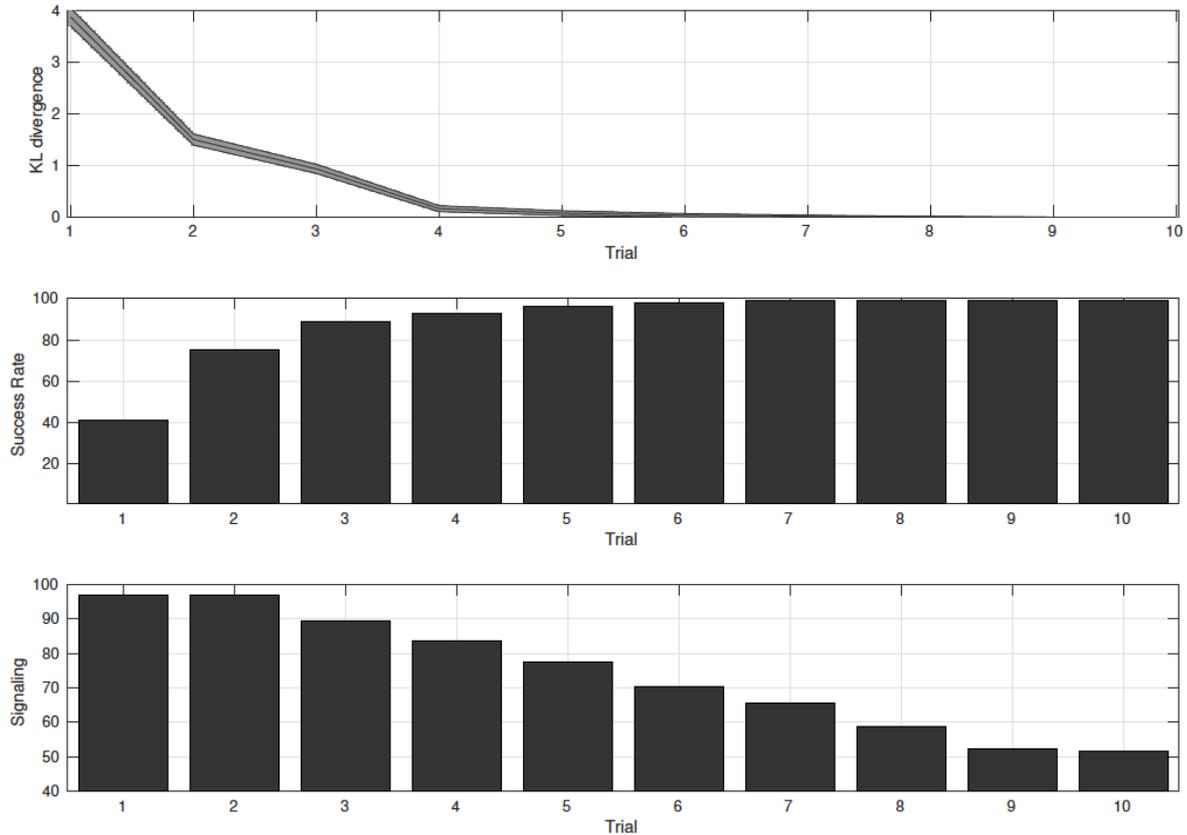

**Fig. 6.** Average results of 100 runs, with the same parameters as Simulation 2 shown in Fig. 7. The format of the first two plots is the same as Fig. 6. The third plot shows the percentage of policies selected by the leader that we label as "social epistemic actions" and prescribe signaling behavior. See the main text for explanation. For example, if the grey agent is the leader, he can select an epistemic policy that passes through L3, L2, L1, L6, L9 and L10 (to reach the red button) or through L3, L4, L5, L8, L13, L12 (to reach the blue button).

Fig. 7 permits appreciating how the leader balances epistemic and pragmatic policies over time. The first panel shows the expected free energy (EFE) (averaged across 100 repetitions) that the leader selects the most useful social epistemic policy (red line) and the most useful pragmatic policy (green line). It shows that the social epistemic policy has a very high probability during the first 5 trials, then its probability decreases until the pragmatic policy becomes the most likely, starting from trial 10. The second panel shows the probability (averaged across 100 repetitions) that the leader selects a social epistemic policy as a function of his uncertainty about the task goal, quantified as the entropy of his belief about the joint task goal. Note that the task goal is a shared representation that encodes both the leader's and the follower's contributions. The entropy over this variable reflects an estimate of the follower's uncertainty, not of the leader's uncertainty (as the leader knows the goal) and decreases over time, as the follower becomes less uncertain.

Notably, the patterns of results shown in Fig. 7B closely corresponds to the findings of a study that uses the "joint maze" setting [50]. Specifically, the study reports that the probability that a (human) participant selects a pragmatic policy is high only when (his or her estimate of) the follower's uncertainty is very low (see Fig. 8A of [50]), which is in good agreement with the pattern shown in our Fig. 7B.

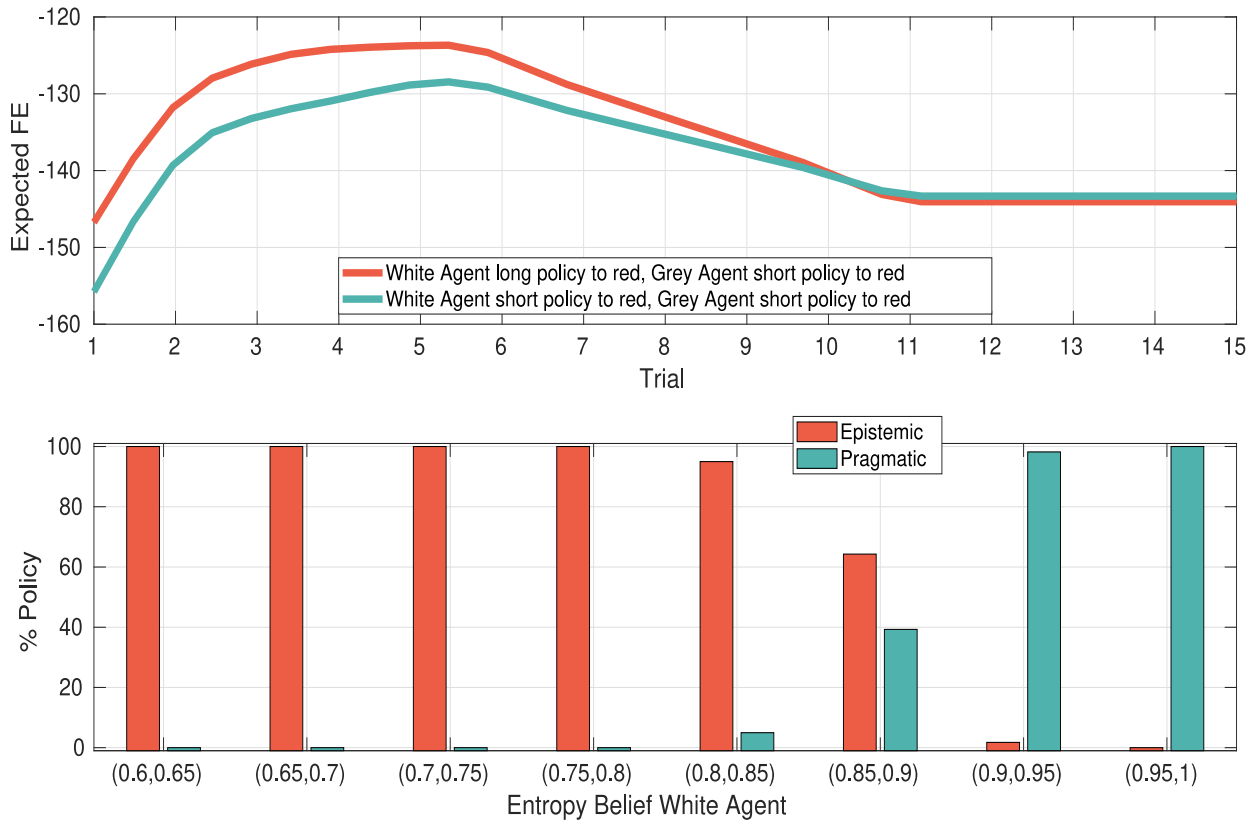

**Fig. 7**. How the leader balances epistemic and pragmatic policies in Simulation 2. Top panel: expected free energy (EFE) of the most useful epistemic (red) and pragmatic policy (green), in the first 10 trials. Bottom panel: frequency of the most useful epistemic policy as a function of the entropy of the leader's belief about the joint task goal. This entropy provides a measure of the (leader's estimate of the) follower's uncertainty.

In sum, Simulation 2 shows that in leader-follower interactions with asymmetric knowledge, leaders select "social epistemic policies" – and therefore sacrifice some efficiency in their choice of movements – to signal their intended goals to the followers and reduce their uncertainty. This signaling behavior is progressively reduced, when the followers become more certain about the joint action goal. This form of signaling was shown in previous computational models that used goal and plan inference, but the models used ad-hoc formulations to promote social epistemic actions [61], [62].

Rather, social epistemic actions emerge naturally from two aspects of our model. The first key aspect of the model that promotes social epistemic behavior is that the expected free energy functional that active inference uses to score policies balances automatically pragmatic and epistemic components of action and policy selection. This means that when uncertainty resolution is necessary, the expected free energy functional automatically promotes epistemic behavior [64]. To illustrate this point, we performed a control simulation (Supplementary Fig. S7) that is the same as Simulation 2, except that we removed the "epistemic component" from the expected free energy that is used to score policies (see the Methods). The results of this control simulation show that the leader selects significantly less social epistemic policies, the behavioral alignment process is slower and the success rate grows more slowly compared to the case in which the full expected free energy is used. This control simulation shows that social epistemic actions are afforded by the expected free energy minimization and that they enhance leader-follower interactions with asymmetric information.

Yet, it is important to remark that in most studies of active inference, epistemic behavior means lowering one's uncertainty, not another's uncertainty, as in our study. The second key aspect of our model that promotes social epistemic behavior – and permits planning how to lower another's uncertainty as opposed to just one's own – is the fact that the leader's generative model includes beliefs about the shared task goal. When scoring his policies (via the expected free energy functional), the leader considers the uncertainty (or the entropy) of the shared task goal; therefore, it assigns high probability to standard "epistemic policies" that lower his own uncertainty (as shown in previous studies) and to "social epistemic policies" that lower the follower's uncertainty [50]. This is important because it shows that active inference agents endowed with shared representations would behave natively in socially-oriented ways, without the need of ad-hoc incentives, beyond joint action optimization.

# V. CONCLUSION

Studies of human-human joint actions have shown some robust trends at both behavioral and neuronal levels, but we still lack a complete formal theory that explains these findings from first principles. Here, we proposed a computational model of *interactive* inference, in which two agents (implemented as active inference agents) coordinate around a joint goal – pressing together either a red or a blue button – that they do not know in advance (Simulation 1) or that only one of them, the leader, knows in advance (Simulation 2).

Our results show that the interactive inference model can successfully reproduce key behavioral and neural signatures of dyadic interactions. Simulation 1 shows that when two agents have the same (uncertain) knowledge about the joint task to be performed, they spontaneously coordinate around a joint goal, align their behavior and task knowledge (here, their beliefs about the joint goal) over time – in keeping with evidence of synchronization of both neuronal activity and kinematics during joint actions [15], [29], [48]. Furthermore, the *interactive* inference is robust to sudden changes of mind of one of the agents, as indexed by the fact that the alignment of behavior and task knowledge is recovered fast. While simple joint tasks such as the "joint maze" that we adopted could be in principle learned by trial and error and without inference, our control simulation (illustrated in Fig. S3 and S4) shows that interactive inference within trials promotes better performance and alignment of behavior and of belief states.

Simulation 2 shows that during dyadic interactions in which knowledge about the joint task goal is asymmetric – specifically, one agent (the "leader") knows the task to be performed but the other agent (the "follower") does not – leaders systematically select "social epistemic policies" in early trials. The social epistemic policies sacrifice some path efficiency to give the follower early cues about the task goal, hence reducing her uncertainty and contributing to optimize the joint action. The results of this simulation are in keeping with a large number of studies of "sensorimotor communication" during dyadic interactions with asymmetric information [24], [50], [58]–[60]. Specifically, our model reproduces two key phenomena of leader-follower interactions. First, in all these tasks, leaders select an apparently less efficient path, which however provides early information about the intended task goal. Second, the selection of these more informative (or social epistemic) policies is dependent on the follower's uncertainty and it is abolished when the follower is (estimated to be) no longer uncertain, as reported in a study that uses our "joint maze" setup [50] and other studies in which the uncertainty of the follower varies across trials [24], [60].

Notably, what renders our model different from previous formalizations is that the leader's social epistemic behavior does not require any ad-hoc mechanism [61], [62]. Rather, social epistemic behavior stems directly from the fact that active inference uses an expected free energy functional that considers epistemic actions on equal ground with pragmatic actions and that the generative model used in our simulation includes shared task knowledge. In other words, active inference agents who cooperate in uncertain conditions and have beliefs about their shared goal can natively select "epistemic" policies that reduce their own uncertainty (as shown in previous simulations [44]) as well as the uncertainty of the other agents (as shown in this simulation). Given that policies of the latter kind – for example, the policies selected by a leader to reduce the follower's uncertainty – are socially oriented, here we call them "social epistemic policies".

Another important feature of our model is its flexibility. Simulations 1 and 2 use exactly the same computational model, except for the fact that in Simulation 2 the "leader" knows the goal, but the follower does not. This implies that active inference is flexible enough to reproduce various aspects of joint action dynamics, without ad-hoc changes of the model. In our simulations, the differences between standard, "leaderless" (Simulation 1) and "leader-follower" (Simulation 2) dynamics emerge as an effect of the strength (and the precision) of the agents' beliefs about the joint goal to be performed. When the agents' beliefs are uncertain, as in Simulation 1, they tend to follow each other to optimize the joint goal – and update (and align) their beliefs afterwards. In this case, the joint outcome (e.g., "red, red" or "blue, blue") can be initially stochastic, but is successively stabilized thanks to the interactive inference. This setting therefore exemplifies a "peer-to-peer" or a "follower-follower" interaction. Yet, it is possible to observe some "leader-follower" dynamics, in the sense that one of the two agents drives the choice of one particular joint task goal. However, in Simulation 1, the role of the leader is not predefined, but rather emerges during the task, as one of the joint goals is stochastically selected during the interaction – and then the two agents stick to it (note however that the situation is different during changes of mind, because the goal is predefined by us rather than being stochastically selected during the interaction).

Rather, Simulation 2 exemplifies the case of a "leader-follower" setup in which the role of the leader is predefined – because the leader has a strong preference for one of the goals. The comparison of Simulations 1 and 2 shows that what defines leaders and followers is simply the strength of the prior belief about the joint task goal (and of its associated outcomes). Our results in Simulation 2 are in keeping with previous active inference models that showed the emergence of behavior synchronization and leader-follower dynamics in joint singing [65] and robotic dyadic interactions [40]. These studies nicely illustrate that several facets of joint actions emerge when two agents infer each other's goals and plans. However, the results reported here go beyond the above studies, by demonstrating the emergence of sensorimotor communication and social epistemic actions when the agents have asymmetric information.

In sum, the simulations illustrated here provide a proof-of-concept that *interactive inference* can reproduce key empirical results of joint action studies. These include, for example, the interactive alignment and synchronization of behavior and neuronal activity (which, in our model, correspond to the belief dynamics) during standard joint actions [15], [29], [48] and the "sensorimotor communication" during dyadic leader-follower joint actions with asymmetric information [24], [50], [58]–[60]. An open objective for future research is extending the empirical validation of this framework by adopting it to model more cases of joint action, beyond the "joint maze" scenario of [50]. Another objective for future research is exploiting this framework to design more effective agents that exploit sensorimotor communication to enhance human-robot joint actions. It has been argued by many researchers that the ease of human-human collaboration rests on our advanced abilities to infer intentions and plans, align representations and select movements that are easily legible and interpretable by our agents. Endowing robots with similar advanced cognitive abilities would permit them to achieve unprecedented levels of success in human-robot collaboration and plausibly increase the trust in robotic agents [43], [66]–[68]. Finally, a key challenge for future research is extending the framework developed here beyond the case of cooperative joint actions, to also cover competitive and mixed cooperative-competitive interactions, which are frequent in multi-agent settings [69].

**Acknowledgement**

This research received funding from the EU Horizon 2020 Framework Programme for Research and Innovation under the Specific Grant Agreement No. 945539 (Human Brain Project SGA3) and the European Research Council under the Grant Agreement No. 820213 (ThinkAhead) to GP and the MUR - PRIN2020 - Grant No. 2020529PCP to FD. The GEFORCE Quadro RTX6000 and Titan GPU cards used for this research were donated by the NVIDIA Corporation. The funders had no role in study design, data collection and analysis, decision to publish, or preparation of the manuscript.

# REFERENCES

[1] A. Dorri, S. S. Kanhere, and R. Jurdak, "Multi-Agent Systems: A Survey," *IEEE Access*, vol. 6, pp. 28573–28593, 2018,

[2] J. Ferber and G. Weiss, *Multi-agent systems: an introduction to distributed artificial intelligence*, vol. 1. Addison-wesley Reading, 1999.

[3] B. Liu, H. Su, R. Li, D. Sun, and W. Hu, "Switching controllability of discrete-time multi-agent systems with multiple leaders and time-delays," *Applied Mathematics and Computation*, vol. 228, pp. 571–588, 2014.

[4] B. Liu, T. Chu, L. Wang, Z. Zuo, G. Chen, and H. Su, "Controllability of switching networks of multi-agent systems," *International Journal of Robust and Nonlinear Control*, vol. 22, no. 6, pp. 630–644, 2012.

[5] S. Su, Z. Lin, and A. Garcia, "Distributed synchronization control of multiagent systems with unknown nonlinearities," *IEEE Transactions on Cybernetics*, vol. 46, no. 1, pp. 325–338, 2015.

[6] B. Anderson, B. Fidan, C. Yu, and D. Walle, "UAV formation control: Theory and application," in *Recent advances in learning and control*, Springer, 2008, pp. 15–33.

[7] N. Krothapalli and A. V. Deshmukh, "Distributed task allocation in multi-agent systems," in *Proceedings of the Institute of Industrial Engineers Annual Conference*, 2002.

[8] W. Ni and D. Cheng, "Leader-following consensus of multi-agent systems under fixed and switching topologies," *Systems & control letters*, vol. 59, no. 3–4, pp. 209–217, 2010.

[9] V. Trianni, D. De Simone, A. Reina, and A. Baronchelli, "Emergence of consensus in a multi-robot network: from abstract models to empirical validation," *IEEE Robotics and Automation Letters*, vol. 1, no. 1, pp. 348–353, 2016.

[10] W. He, B. Xu, Q.-L. Han, and F. Qian, "Adaptive Consensus Control of Linear Multiagent Systems With Dynamic Event-Triggered Strategies," *IEEE Transactions on Cybernetics*, vol. 50, no. 7, pp. 2996–3008, 2020

[11] C. Castelfranchi, "Modeling social interaction for AI agents," in *IJCAI-97. Proceedings of the Fifteenth International Joint Conference on Artificial Intelligence*, 1997, pp. 1567–1576.

[12] R. Conte and C. Castelfranchi, *Cognitive and Social Action*. London, UK: University College London, 1995.

[13] C. Castelfranchi and R. Falcone, *Trust Theory: A socio-cognitive and computational model*. Wiley, 2010.

[14] R. Sun, "Cognitive science meets multi-agent systems: A prolegomenon," *Philosophical psychology*, vol. 14, no. 1, pp. 5–28, 2001.

[15] N. Sebanz, H. Bekkering, and G. Knoblich, "Joint action: bodies and minds moving together.," *Trends Cogn Sci*, vol. 10, no. 2, pp. 70–76, 2006

[16] G. Pezzulo, F. Donnarumma, H. Dindo, A. D'Ausilio, I. Konvalinka, and C. Castelfranchi, "The body talks: Sensorimotor communication and its brain and kinematic signatures," *Physics of Life Reviews*, Jun. 2018, doi: 10.1016/j.plrev.2018.06.014.

[17] O. Oullier and J. A. Kelso, "Social coordination, from the perspective of coordination dynamics," in *Encyclopedia of complexity and systems science*, Springer, 2009, pp. 8198–8213. Accessed: Feb. 23, 2016

[18] G. Knoblich and N. Sebanz, "Evolving intentions for social interaction: from entrainment to joint action.," *Philos Trans R Soc Lond B Biol Sci*, vol. 363, no. 1499, pp. 2021–2031, Jun. 2008,

[19] G. Pezzulo, F. Donnarumma, S. Ferrari-Toniolo, P. Cisek, and A. Battaglia-Mayer, "Shared population-level dynamics in monkey premotor cortex during solo action, joint action and action observation," *Progress in Neurobiology*, vol. 210, p. 102214, Mar. 2022,

[20] G. Pezzulo, P. Iodice, F. Donnarumma, H. Dindo, and G. Knoblich, "Avoiding accidents at the champagne reception: A study of joint lifting and balancing," *Psychological Science*, 2017.

[21] G. Pezzulo, L. Roche, and L. Saint-Bauzel, "Haptic communication optimises joint decisions and affords implicit confidence sharing," *Sci Rep*, vol. 11, no. 1, p. 1051, Jan. 2021

[22] G. Knoblich and J. S. Jordan, "Action coordination in groups and individuals: learning anticipatory control.," *J Exp Psychol Learn Mem Cogn*, vol. 29, no. 5, pp. 1006–1016, Sep. 2003, doi: 10.1037/0278-7393.29.5.1006.

[23] R. P. R. D. van der Wel, C. Becchio, A. Curioni, and T. Wolf, "Understanding joint action: Current theoretical and empirical approaches," *Acta Psychologica*, vol. 215, p. 103285, Apr. 2021

[24] M. Candidi, A. Curioni, F. Donnarumma, L. M. Sacheli, and G. Pezzulo, "Interactional leader–follower sensorimotor communication strategies during repetitive joint actions," *Journal of The Royal Society Interface*, vol. 12, no. 110, p. 20150644, Sep. 2015

[25] F. Visco-Comandini *et al.*, "Do non-human primates cooperate? Evidences of motor coordination during a joint action task in macaque monkeys," *Cortex*, vol. 70, pp. 115–127, 2015.

[26] C. Becchio, A. Koul, C. Ansuini, C. Bertone, and A. Cavallo, "Seeing mental states: An experimental strategy for measuring the observability of other minds," *Phys Life Rev*, vol. 24, pp. 67–80, Mar. 2018

[27] M. Coco *et al.*, "Multilevel behavioral synchronisation in a joint tower-building task," *IEEE Transactions on Cognitive and Developmental Systems*, vol. PP, no. 99, pp. 1–1, 2016

[28] A. D'Ausilio, G. Novembre, L. Fadiga, and P. E. Keller, "What can music tell us about social interaction?," *Trends Cogn. Sci. (Regul. Ed.)*, vol. 19, no. 3, pp. 111–114, Mar. 2015

[29] P. E. Keller, G. Novembre, and M. J. Hove, "Rhythm in joint action: psychological and neurophysiological mechanisms for real-time interpersonal coordination," *Phil. Trans. R. Soc. B*, vol. 369, no. 1658, p. 20130394, 2014

[30] D. M. Wolpert, K. Doya, and M. Kawato, "A unifying computational framework for motor control and social interaction.," *Philos Trans R Soc Lond B Biol Sci*, vol. 358, no. 1431, pp. 593–602, Mar. 2003, doi: 10.1098/rstb.2002.1238.

[31] H. Dindo, D. Zambuto, and G. Pezzulo, "Motor simulation via coupled internal models using sequential Monte Carlo," in *Proceedings of IJCAI 2011*, 2011, pp. 2113–2119.

[32] H. Dindo, F. Donnarumma, F. Chersi, and G. Pezzulo, "The intentional stance as structure learning: a computational perspective on mindreading," *Biological cybernetics*, vol. 109, no. 4–5, pp. 453–467, 2015.

[33] G. Pezzulo and H. Dindo, "What should I do next? Using shared representations to solve interaction problems," *Experimental Brain Research*, vol. 211, no. 3, pp. 613–630, 2011.

[34] G. Pezzulo, F. Donnarumma, and H. Dindo, "Human Sensorimotor Communication: A Theory of Signaling in Online Social Interactions," *PLoS ONE*, vol. 8, no. 11, p. e79876, Nov. 2013

[35] C. L. Baker, R. Saxe, and J. B. Tenenbaum, "Action understanding as inverse planning," *Cognition*, vol. 113, no. 3, pp. 329–349, Sep. 2009

[36] T. D. Ullman *et al.*, "Help or hinder: Bayesian models of social goal inference," in *Proceedings of Neural Information Processing Systems*, 2009.

[37] N. Tang, S. Stacy, M. Zhao, G. Marquez, and T. Gao, "Bootstrapping an Imagined We for Cooperation.," in *CogSci*, 2020.

[38] J. Hwang, J. Kim, A. Ahmadi, M. Choi, and J. Tani, "Dealing with large-scale spatio-temporal patterns in imitative interaction between a robot and a human by using the predictive coding framework," *IEEE Transactions on Systems, Man, and Cybernetics: Systems*, vol. 50, no. 5, pp. 1918–1931, 2018.

[39] J. Tani, R. Nishimoto, J. Namikawa, and M. Ito, "Codevelopmental learning between human and humanoid robot using a dynamic neural-network model," *IEEE Trans Syst Man Cybern*, vol. 38, no. 1, pp. 43–59, 2008

[40] N. Wirkuttis and J. Tani, "Leading or Following? Dyadic Robot Imitative Interaction Using the Active Inference Framework," *IEEE Robotics and Automation Letters*, vol. 6, no. 3, pp. 6024–6031, Jul. 2021

[41] K. Dautenhahn *et al.*, "How may I serve you? A robot companion approaching a seated person in a helping context," in *Proc of the 1st ACM SIGCHI/SIGART conference on Human-robot interaction*, 2006, pp. 172–179.

[42] A. Clodic and R. Alami, "What Is It to Implement a Human-Robot Joint Action?," in *Robotics, AI, and Humanity*, Springer, Cham, 2021, pp. 229–238.

[43] A. Clodic, E. Pacherie, R. Alami, and R. Chatila, "Key Elements for Human Robot Joint Action," p. 159, 2017.

[44] T. Parr, G. Pezzulo, and K. J. Friston, *Active Inference: The Free Energy Principle in Mind, Brain, and Behavior*. MIT Press, 2022.

[45] Q. Song, F. Liu, H. Su, and A. V. Vasilakos, "Semi-global and global containment control of multi-agent systems with second-order dynamics and input saturation," *International Journal of Robust and Nonlinear Control*, vol. 26, no. 16, pp. 3460–3480, 2016.

[46] I. Konvalinka, P. Vuust, A. Roepstorff, and C. D. Frith, "Follow you, follow me: continuous mutual prediction and adaptation in joint tapping.," *Q J Exp Psychol (Colchester)*, vol. 63, no. 11, pp. 2220–2230, Nov. 2010

[47] J. C. Skewes, L. Skewes, J. Michael, and I. Konvalinka, "Synchronised and complementary coordination mechanisms in an asymmetric joint aiming task," *Experimental brain research*, vol. 233, no. 2, pp. 551–565, 2015.

[48] G. Novembre, G. Knoblich, L. Dunne, and P. E. Keller, "Interpersonal synchrony enhanced through 20 Hz phase-coupled dual brain stimulation," *Soc Cogn Affect Neurosci*, Jan. 2017

[49] S. Garrod and M. J. Pickering, "Joint Action, Interactive Alignment, and Dialog," *Topics in Cognitive Science*, vol. 1, no. 2, pp. 292–304, 2009

[50] G. Pezzulo and H. Dindo, "What should I do next? Using shared representations to solve interaction problems," *Experimental Brain Research*, vol. 211, no. 3, pp. 613–630, 2011.

[51] C. Bishop, *Pattern Recognition and Machine Learning*. Springer, 2006.

[52] D. Maisto, F. Donnarumma, and G. Pezzulo, "Nonparametric Problem-Space Clustering: Learning Efficient Codes for Cognitive Control Tasks," *Entropy*, vol. 18, no. 2, p. 61, Feb. 2016,

[53] K. Von Fritz, "ΝΟΟΣ and Noein in the Homeric Poems," *Classical Philology*, vol. 38, no. 2, pp. 79–93, 1943.

[54] G. Gergely and G. Csibra, "Teleological reasoning in infancy: the naive theory of rational action," *Trends in Cognitive Sciences*, vol. 7, pp. 287–292, 2003.

[55] N. Sebanz, G. Knoblich, W. Prinz, and E. Wascher, "Twin peaks: an ERP study of action planning and control in co-acting individuals.," *J Cogn Neurosci*, vol. 18, no. 5, pp. 859–870, May 2006

[56] N. Sebanz, G. Knoblich, and W. Prinz, "How two share a task: corepresenting stimulus-response mappings.," *J Exp Psychol Hum Percept Perform*, vol. 31, no. 6, pp. 1234–1246, Dec. 2005

[57] G. Knoblich and N. Sebanz, "Evolving intentions for social interaction: from entrainment to joint action.," *Philos Trans R Soc Lond B Biol Sci*, vol. 363, no. 1499, pp. 2021–2031, Jun. 2008

[58] L. M. Sacheli, E. Tidoni, E. F. Pavone, S. M. Aglioti, and M. Candidi, "Kinematics fingerprints of leader and follower role-taking during cooperative joint actions," *Exp Brain Res*, vol. 226, no. 4, pp. 473–486, May 2013

[59] C. Vesper and M. J. Richardson, "Strategic communication and behavioral coupling in asymmetric joint action," *Exp Brain Res*, May 2014, doi: 10.1007/s00221-014-3982-1.

[60] F. Leibfried, J. Grau-Moya, and D. A. Braun, "Signaling equilibria in sensorimotor interactions," *Cognition*, vol. 141, pp. 73–86, Aug. 2015,

[61] G. Pezzulo, F. Donnarumma, and H. Dindo, "Human Sensorimotor Communication: A Theory of Signaling in Online Social Interactions," *PLoS ONE*, vol. 8, no. 11, p. e79876, Nov. 2013

[62] G. Pezzulo, F. Donnarumma, H. Dindo, A. D'Ausilio, I. Konvalinka, and C. Castelfranchi, "The body talks: Sensorimotor communication and its brain and kinematic signatures," *Physics of life reviews*, 2018.

[63] K. Friston, T. FitzGerald, F. Rigoli, P. Schwartenbeck, and G. Pezzulo, "Active Inference: A Process Theory," *Neural Comput*, vol. 29, no. 1, pp. 1–49, Jan. 2017, doi: 10.1162/NECO_a_00912.

[64] T. Parr, G. Pezzulo, and K. J. Friston, *Active Inference: The Free Energy Principle in Mind, Brain, and Behavior*. MIT Press, 2022.

[65] K. Friston and C. Frith, "A Duet for one," *Consciousness and cognition*, vol. 36, pp. 390–405, 2015.

[66] K. Belhassein *et al.*, "Addressing joint action challenges in HRI: Insights from psychology and philosophy," *Acta Psych*, vol. 222, p. 103476, 2022

[67] F. Donnarumma, H. Dindo, and G. Pezzulo, "Sensorimotor communication for humans and robots: improving interactive skills by sending coordination signals," *IEEE Transactions on Cognitive and Developmental Systems*, vol. PP, no. 99, pp. 1–1, 2017

[68] A. D. Dragan, K. C. Lee, and S. S. Srinivasa, "Legibility and predictability of robot motion," in *2013 8th ACM/IEEE International Conference on Human-Robot Interaction (HRI)*, 2013, pp. 301–308.

[69] T. T. Nguyen, N. D. Nguyen, and S. Nahavandi, "Deep Reinforcement Learning for Multiagent Systems: A Review of Challenges, Solutions, and Applications," *IEEE Transactions*

# Supplementary materials

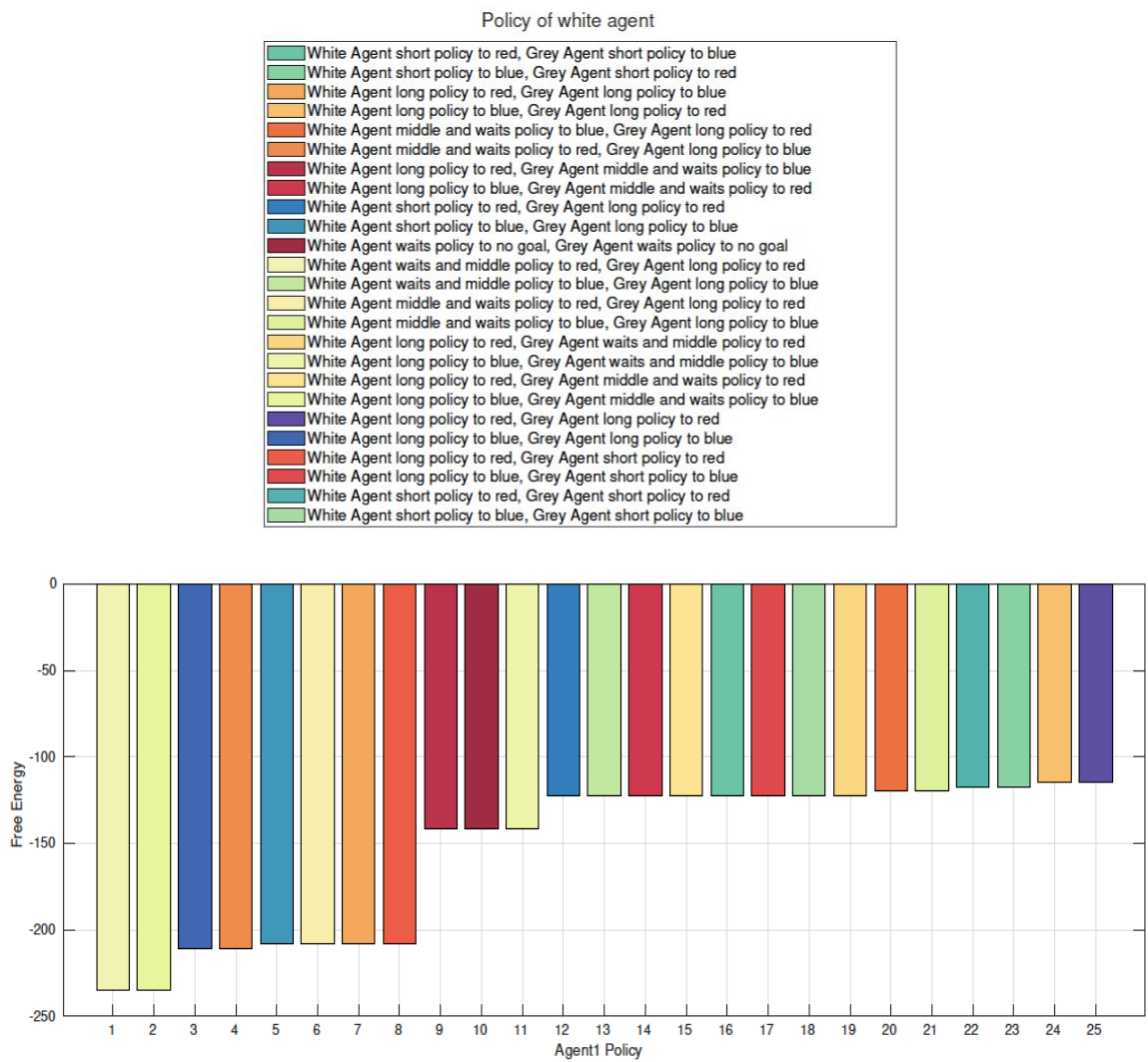

**Fig. S1.** Score (expected free energy) that the grey agent assigns to its 25 policies at the beginning of the first trial of Simulation 1. The policies are ordered according to their expected free energy; please note that the policies having the highest expected free energy (to the left) are the least preferred, whereas those having the lowest expected free energy (to the right) are the most preferred. The two equally preferred policies (those having the lowest expected free energy) are the 24th and the 25th, which infers that both the grey agent (T1) and the white agent (T2) will go straight to the same button, red or blue. Here, "going straight" means passing through L7 and L11 (for the grey agent) passing through L15 and L11 (for the white agent). Rather, "long policy" means (for example, for the grey agent) going through L4, L5, L8 and L13 to reach the blue button. Please note also that the two most dis-preferred policies (those having the highest expected free energy) are the 1st and 2nd, which infer that the two agents will go directly to the two opposite goals.

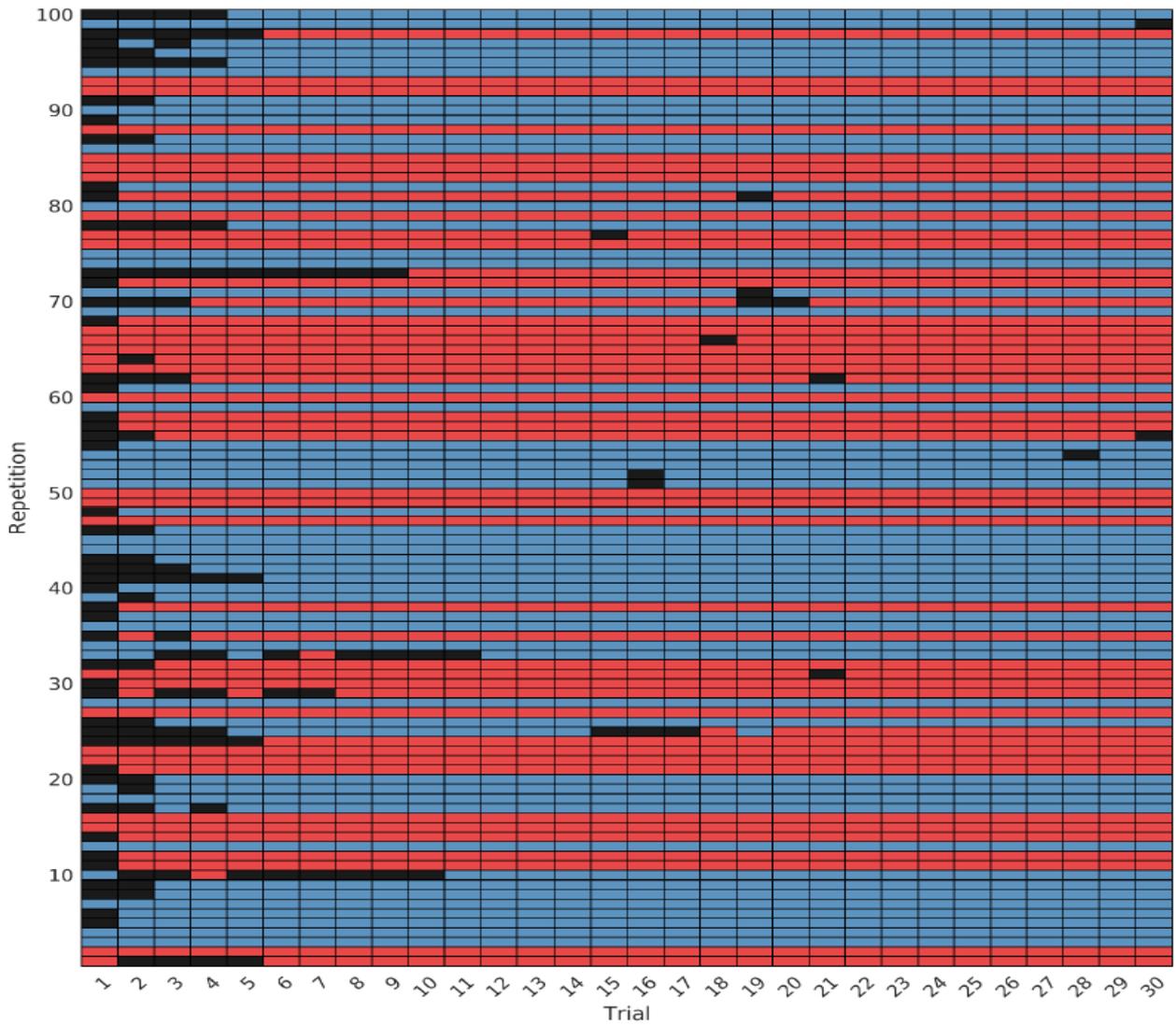

**Fig. S2.** Results of 100 repetitions of Simulation 1, for 30 trials and without changes of mind. The blue and red squares indicate that the agents successfully solved the task by pressing the blue and red buttons, respectively. The black squares indicate failures.

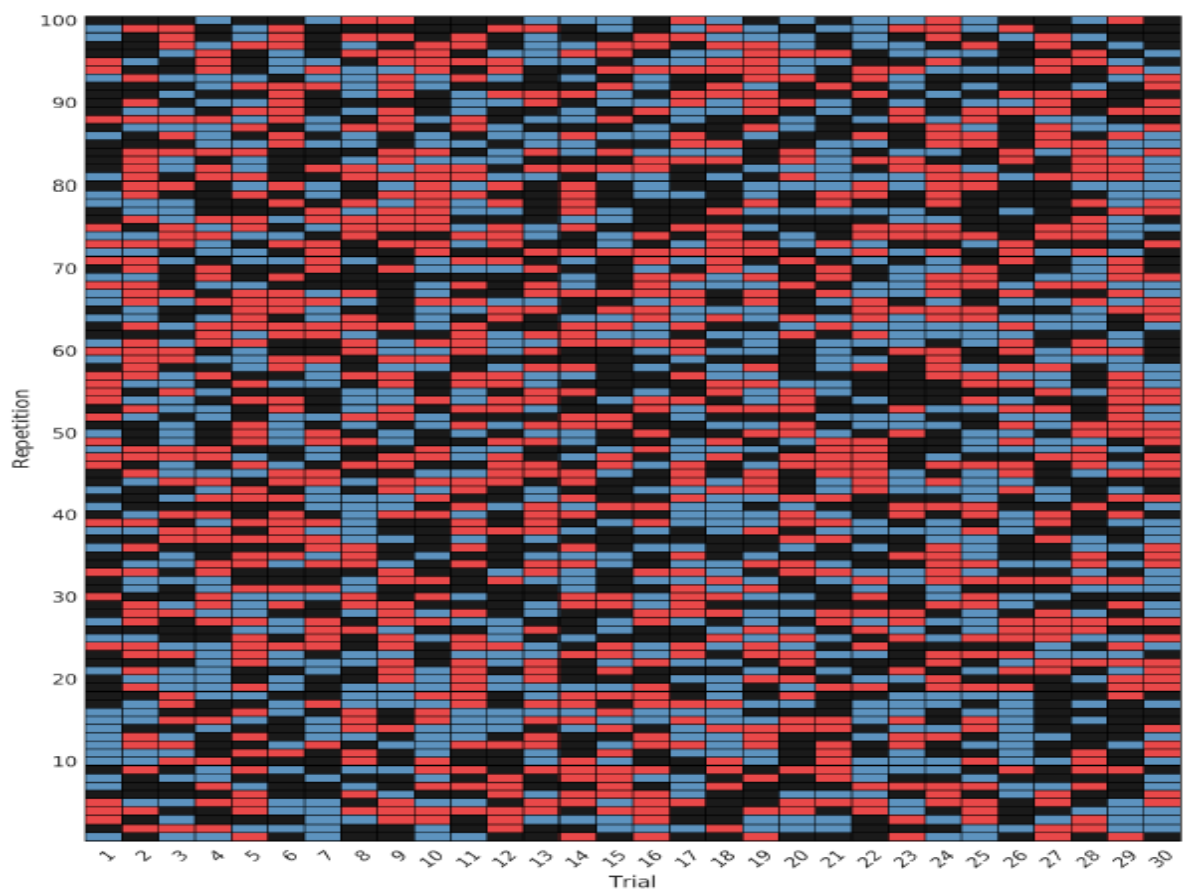

**Fig. S3.** Results of 100 repetitions of Simulation 1, for 30 trials and without changes of mind, when interactive inference is prevented. The blue and red squares indicate that the agents successfully solved the task by pressing the blue and red buttons, respectively. The black squares indicate failures. See the main text for explanation.

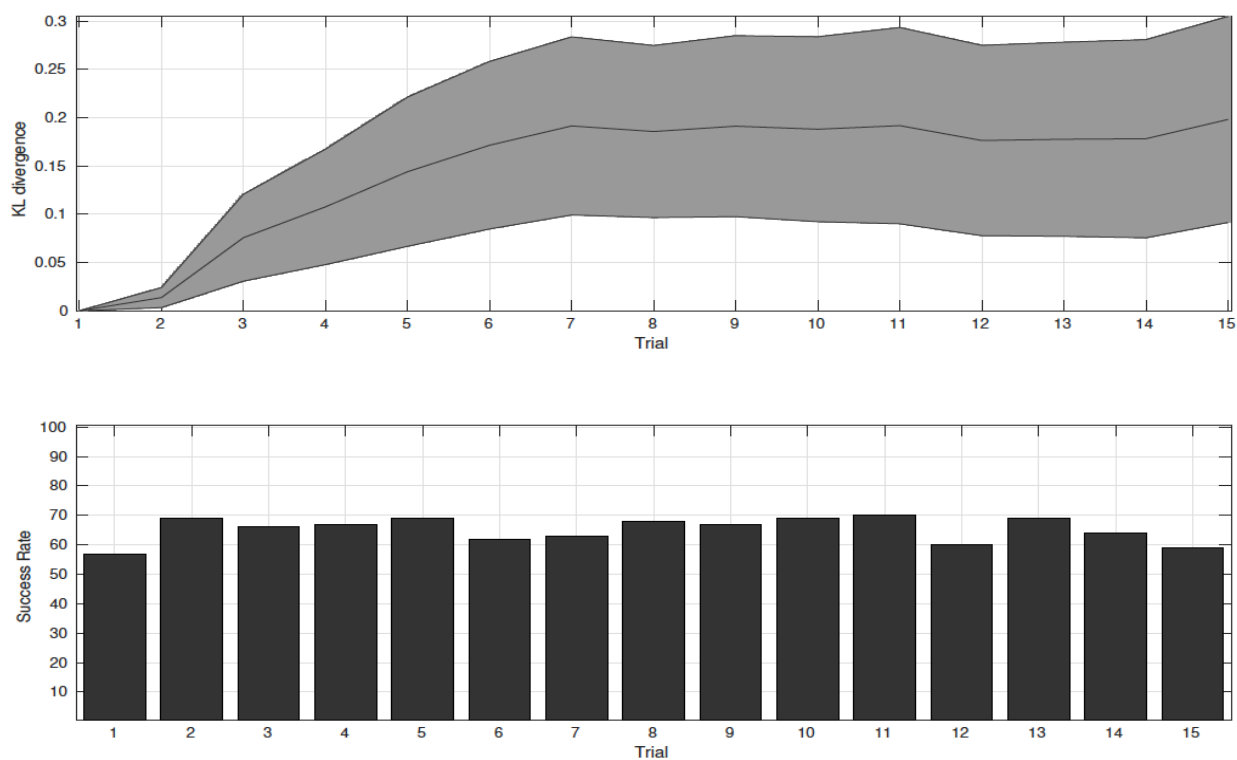

**Fig. S4.** Average results of 100 runs, with the same parameters as Simulation 1, for 15 trials, when interactive inference is prevented. The top panel shows a measure of belief alignment of the agents: the KL divergence between their beliefs about task goals. The mean value is in black and the standard deviation of the mean is in grey. The bottom panel shows a histogram of mean success rate.

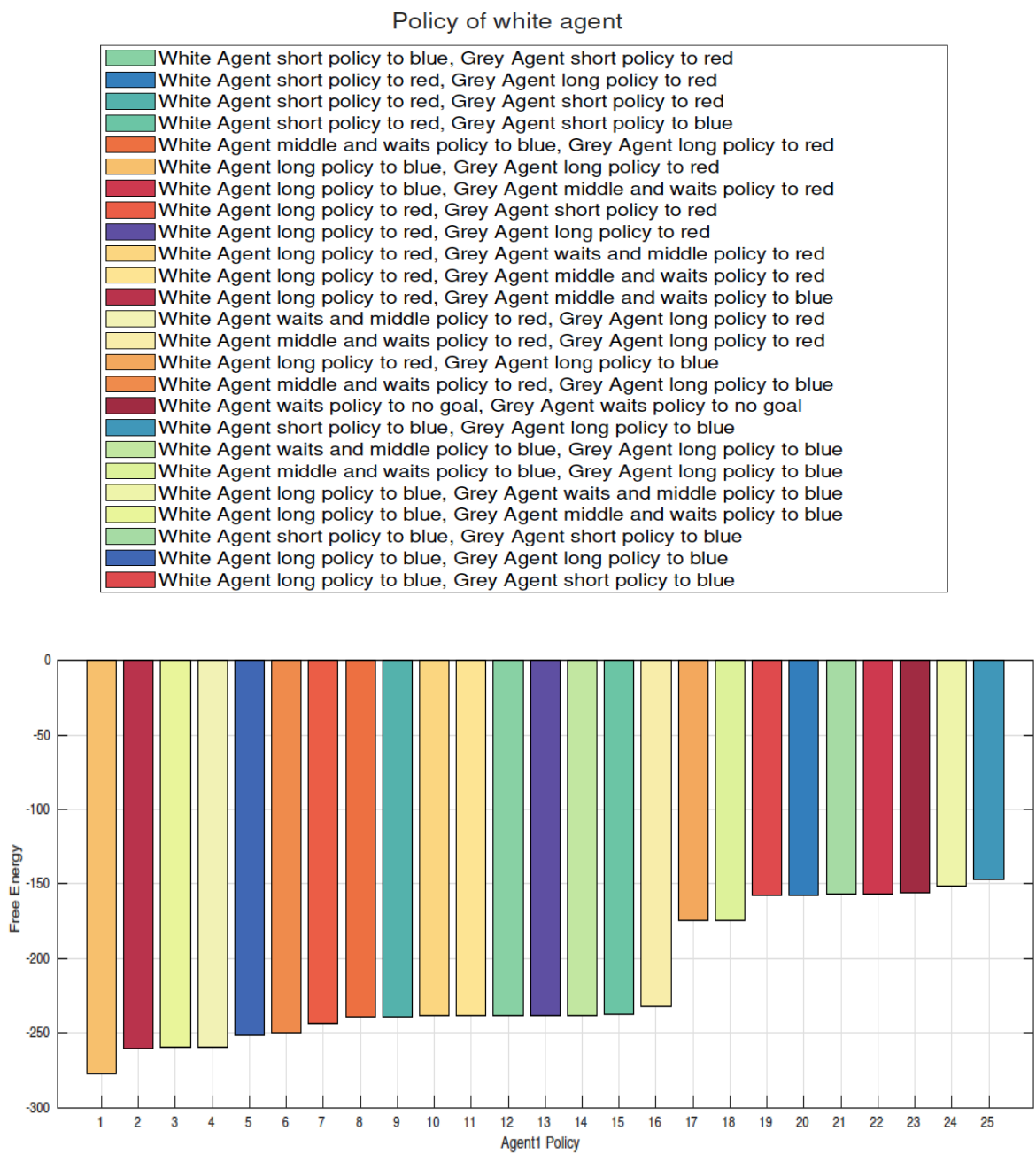

**Fig. S5.** Score (expected free energy) that the grey agent – acting as the leader – assigns to its 25 policies at the beginning of the first trial of Simulation 2. The policies are ordered according to their expected free energy; please note that the policies having the highest expected free energy (to the left) are the least preferred, whereas those having the lowest expected free energy (to the right) are the most preferred. The most preferred policy, the 25th, infers that the leader will follow the long path to the red goal and the follower will follow the short path to the red goal. The second most preferred policy, the 25th, infers that both agents will follow the long path to the red goal. These (and most of the other preferred) policies are "social epistemic policies" for the leader.

**Fig. S6.** Results of 100 repetitions of Simulation 2, for 10 trials. The red squares indicate that the agents successfully solved the task by pressing the red buttons. The black squares indicate failures.

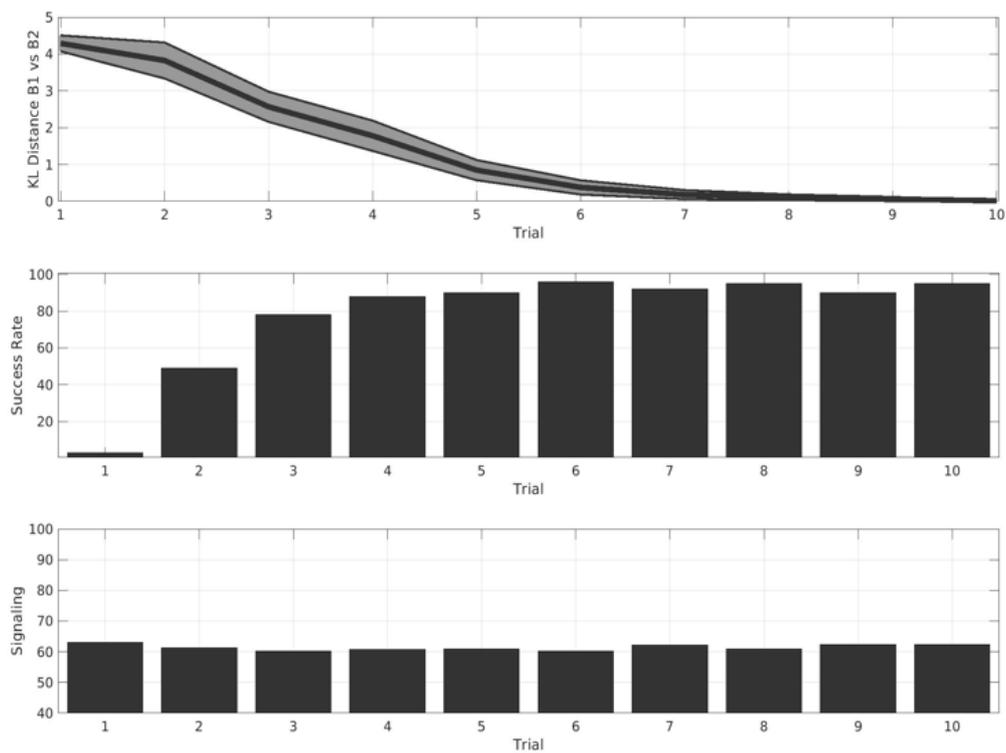

**Fig. S7.** Average results of 100 runs, with the same parameters as Simulation 2, but after removing the "epistemic component" from the expected free energy Fig.. See the main text for explanation.

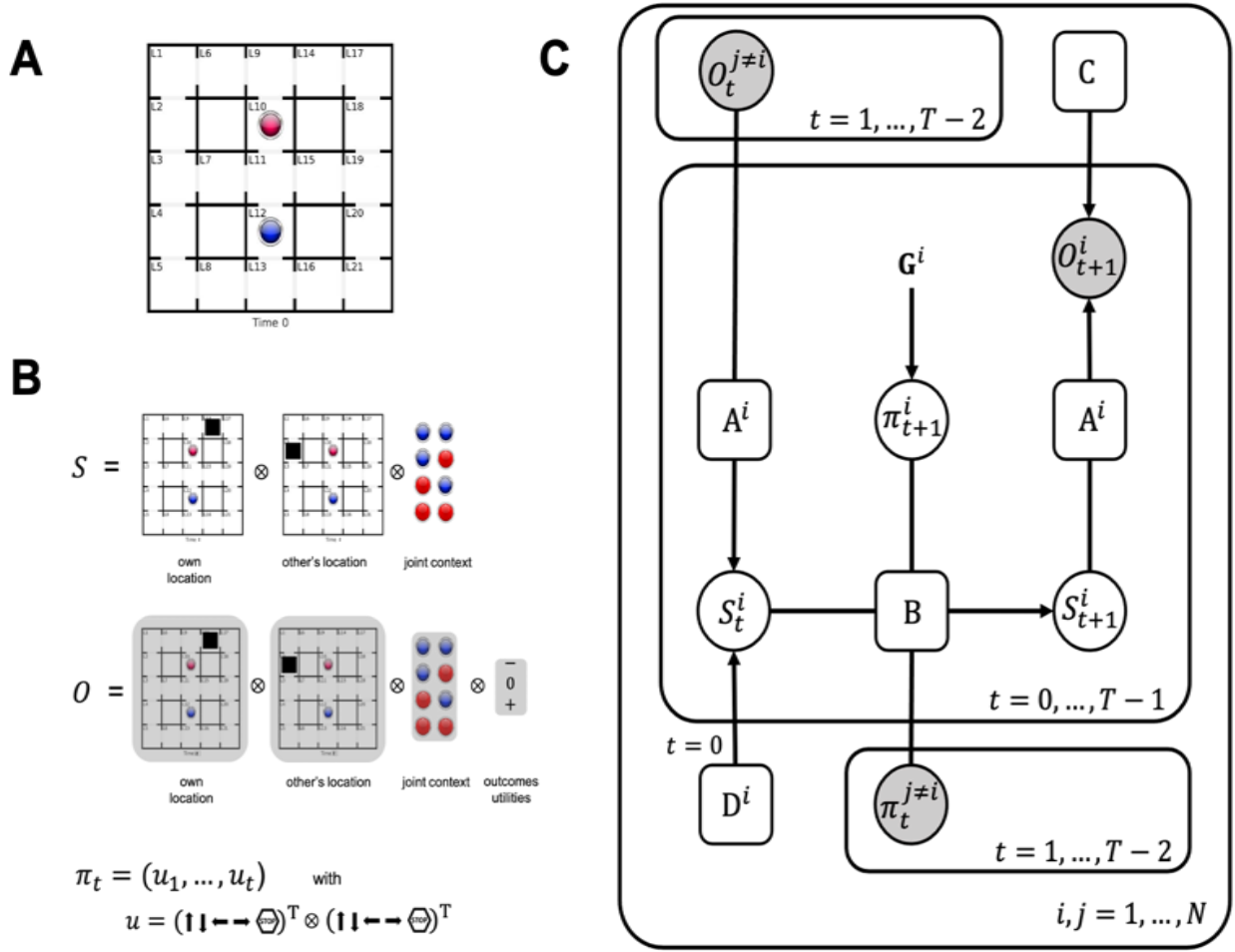

**Fig. S8**. Schematic illustration of the multi-agent active inference model used in the current article. (A) Sketch of the "joint maze" task illustrated in Fig. 1. Two agents, "grey" ("i") and "white" ("j"), start from the locations L3 and L19 and their goal is to reach either the red (L10) or the blue (L12) goal locations. (B) The stochastic variables of the generative model adopted for the agents. $S^i$, the hidden states, are the tensorial product between the position of the ith agent, the position of the jth agent and the joint goal context belief. $O^i$ is the observation tensor defined as the product between the observed position of the ith agent, the observed position of the jth agent, the observed joint context, and the outcome utility. The symbol $u_t$ denotes the joint action at the time t defined as the tensorial product between the action of the ith agent and the action of the jth agent. A sequence of joint actions $(u_1, u_2, ..., u_t)$ forms a policy, indicated by the Greek letter $\pi_t$. (C) A probabilistic graph of the generative model for multi-agent active inference. The plate notation indicates that the structures held in a box vary as functions of the specified indexes. Therefore, each agent denoted with an index (from 1 to N) of the outer box executes the process in the inner box, whose time horizon is t = $0, ..., T - 1$. During the execution of that process, each agent sends his position and last action and receives the positions and the actions of the other agents. Each agent uses the information from the other agents to infer the evolution of the entire scenario and to update his own model. Note that this scheme could be extended to multiple agents. Please see the main text for details.

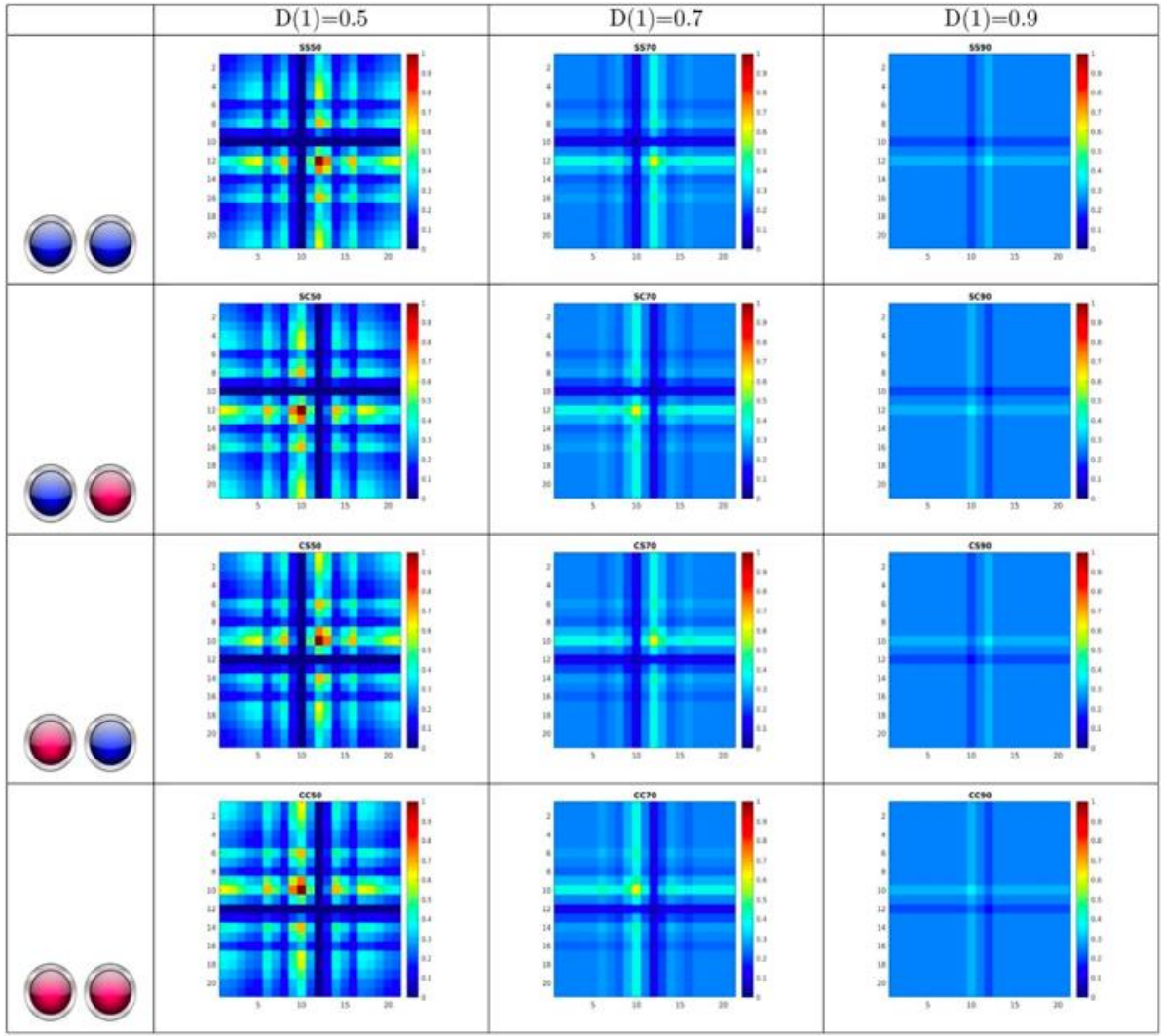

**Fig. S9.** Schematic illustration of the tensor $A_3^i$, playing the role of a "salience map", which varies as a function of the agent's beliefs about the joint task goal. The figure is organized as a table, in which the columns correspond to different values of the mode of the initial belief about the task goal i.e., $\max(D_3^i)$) and the rows correspond to the four possible joint goals: "blue, blue", "blue, red", "red, blue", "red, red". Each of the 12 matrices shows a specific tensor $A_3^i$, which encodes the (color-coded) salience of the positions of the two agents in the maze. As illustrated in the figure, the higher the initial belief, the less informative the "salience map". See the main text for explanation.